\def\paperTitle{SDGBiasBench: Benchmarking and Mitigating Vision--Language Models' Biases in Sustainable Development Goals}

\def\authorBlock{
{
    Zihang Lin\,\textnormal{\textsuperscript{1,2,*}}
    \ \
    Huaiyuan Qin\,\textnormal{\textsuperscript{2,$\dagger$}}
    \ \
    Muli Yang\,\textnormal{\textsuperscript{2}}
    \ \
    Hongyuan Zhu\,\textnormal{\textsuperscript{2,\Letter}}
} \\
\\
{
\fontsize{10.9}{16}\selectfont
\textsuperscript{1}Nanyang Technological University
}\\
{
\fontsize{10.9}{16}\selectfont
\textsuperscript{2}Institute of Advanced Intelligence and Computing (IAIC), A*STAR, Singapore
}\\
{   
\fontsize{10.9}{16}\selectfont
    \textsuperscript{$\dagger$}Project Lead. 
    \ \ 
    \textsuperscript{\Letter}Corresponding Author.
} \\ 

{\tt
\small{linz0071@e.ntu.edu.sg, \{qinhy, yangml, zhuh\}@a-star.edu.sg}
}
}

\newif\ifreview
\newif\ifarxiv
\newif\iffinal

\newcommand{\arxiv}{\arxivtrue \reviewfalse \finalfalse}

\arxiv

\pdfoutput=1
\documentclass[11pt]{article}
\ifreview \usepackage[review]{meta/acl} \fi
\ifarxiv \usepackage[preprint]{meta/acl} \fi
\iffinal \usepackage{meta/acl} \fi

\usepackage{times}
\usepackage{latexsym}

\usepackage[T1]{fontenc}

\usepackage[utf8]{inputenc}

\usepackage{microtype}

\usepackage{inconsolata}

\usepackage{graphicx}

\usepackage{fix-cm}
\usepackage{array}
\usepackage{bbm}
\usepackage{nicematrix}

\usepackage{tipa}
\usepackage{dsfont}
\usepackage{etoolbox}

\usepackage[noend]{algorithmic}
\usepackage{algorithm}

\usepackage{float}
\usepackage{newfloat}
\usepackage{listings}
\floatstyle{ruled}
\newfloat{listing}{tb}{lst}{}
\floatname{listing}{\small Algorithm}

\definecolor{mygray}{RGB}{234,234,234}

\usepackage{adjustbox}

\definecolor{darkgreen}{rgb}{0.13, 0.55, 0.13}

\usepackage{amsmath}	
\usepackage{amssymb}	
\usepackage{booktabs}
\usepackage{epsfig}
\usepackage{caption}
\usepackage{float}
\usepackage{placeins}
\usepackage{color, colortbl}
\usepackage{stfloats}
\usepackage{enumitem}
\setlist{leftmargin=5.5mm}

\usepackage{tabularx}
\usepackage{xstring}
\usepackage{multirow}
\usepackage{xspace}
\usepackage{url}
\usepackage{subcaption}

\ifarxiv  \fi

\newcommand{\R}[1]{{%
    \textbf{%
        \ifstrequal{#1}{1}{\textcolor{red}{R#1}}{%
        \ifstrequal{#1}{2}{\textcolor{blue}{R#1}}{%
        \ifstrequal{#1}{3}{\textcolor{magenta}{R#1}}{%
        \ifstrequal{#1}{4}{\textcolor{teal}{R#1}}{%
                           \textcolor{cyan}{R#1}%
        }}}}%
    }%
}}

\definecolor{Gray}{gray}{0.5}
\definecolor{nicergreen}{rgb}{0.13, 0.54, 0.13}
\definecolor{nicered}{rgb}{0.83, 0.16, 0.16}
\definecolor{Highlight}{HTML}{39b54a}

\newcommand{\cgaphl}[2]{
\fontsize{6pt}{1em}\selectfont{\textcolor{nicergreen}{\textsuperscript{${#1}$\textbf{#2}}}}
}

\usepackage{appendix}
\usepackage{footnote}

\usepackage{tcolorbox}
\newtcolorbox[auto counter]{conclusions}[1][]{
  title={\bfseries Finding \thetcbcounter},
  coltitle=white,
  #1
}

\makeatletter
\DeclareRobustCommand\onedot{\futurelet\@let@token\@onedot}
\def\@onedot{\ifx\@let@token.\else.\null\fi\xspace}

\def\eg{\emph{e.g}\onedot} 
\def\ie{\emph{i.e}\onedot} 
 
 \def\vs{\emph{vs}\onedot}

\makeatother

\usepackage{ifsym, marvosym}

\let\svthefootnote\thefootnote
\newcommand\freefootnote[1]{%
  \let\thefootnote\relax%
  \footnotetext{#1}%
  \let\thefootnote\svthefootnote%
}

\usepackage{tocloft}
\usepackage{cuted}
\usepackage{setspace}
\usepackage{lineno}

\usepackage{xr-hyper}

\makeatletter
\newcommand*{\addFileDependency}[1]{
  \typeout{(#1)}
  \@addtofilelist{#1}
  \IfFileExists{#1}{}{\typeout{No file #1.}}
}

\makeatother

\usepackage[capitalize]{cleveref}
\crefname{section}{Sec.}{Secs.}
\crefname{table}{Table}{Tables}
\crefname{figure}{Fig.}{Figs.}

\ifarxiv \crefname{appendix}{App.}{Apps.}
\else \crefname{appendix}{Suppl.}{Suppls.} \fi

\begin{document}
\title{\paperTitle}
\author{\authorBlock}

\twocolumn[{
\renewcommand\twocolumn[1][]{#1}
\maketitle
\begin{center}
    \vspace{-8mm}
    \setlength{\abovecaptionskip}{6pt}
    \centering
    \includegraphics[width= 0.8\textwidth]{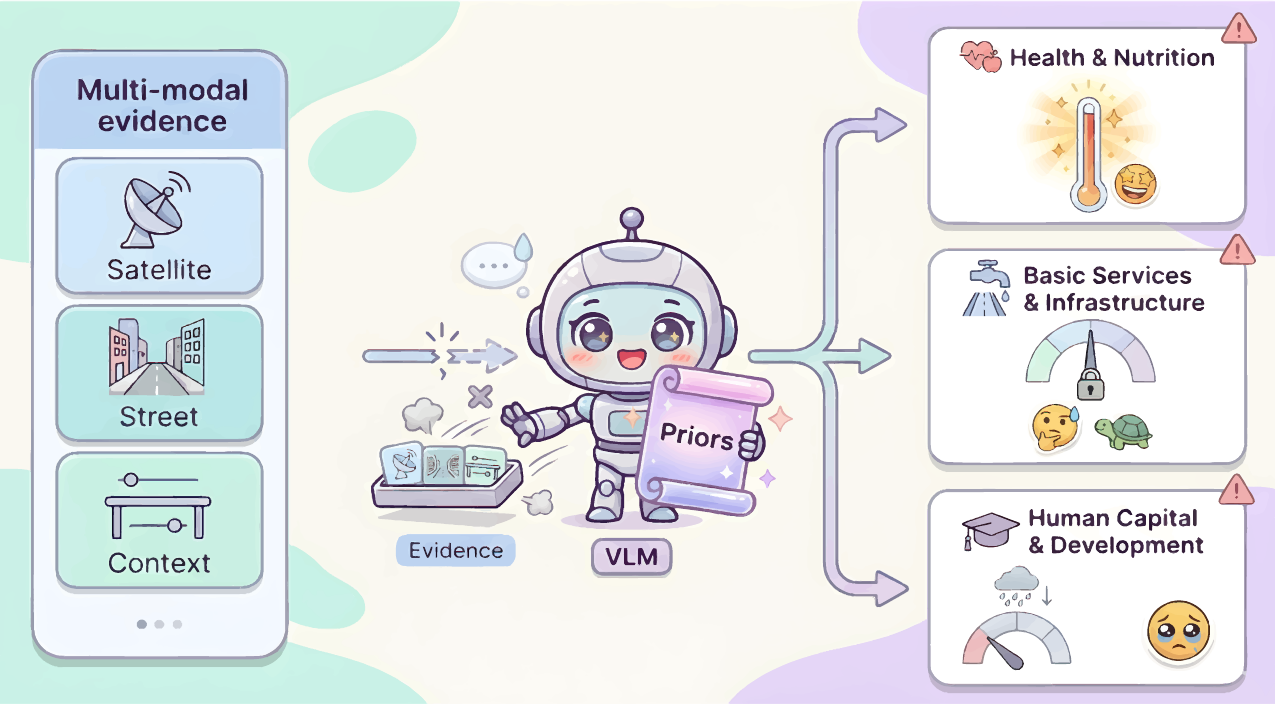}
    \captionof{figure}{Sustainable Development Goals (SDGs) monitoring demands multi-step reasoning over satellite imagery, street-level cues, and structured context.
    However, VLMs always tend to under-use or mis-integrate this evidence and instead rely on SDG-specific priors, yielding directionally skewed predictions across SDG domains. 
    We therefore introduce \textit{\textbf{SDGBiasBench}} to expose decision- and estimation-level SDG bias under controlled evidence availability.
    }
    \label{fig:figure1}
    \vspace{0.5cm}
\end{center}

}]

\ifreview
\else
\freefootnote{\hspace{-15pt}\textsuperscript{*}This paper was completely accomplished when Zihang Lin interned under Dr. Hongyuan Zhu's supervision at A*STAR.}
\fi

\begin{abstract}

Assessing progress toward the Sustainable Development Goals (SDGs) requires multi-step reasoning over visual cues, contextual knowledge, and development indicators, where incomplete evidence use and imperfect evidence integration can introduce hidden prediction biases. Real-world SDG monitoring further spans both qualitative judgments and quantitative estimation. However, existing benchmarks typically evaluate these aspects in isolation, obscuring systematic biases that emerge when models substitute priors for evidence. To address this gap, we propose \textbf{SDGBiasBench}, a large-scale benchmark suite for SDG-oriented vision–language reasoning. Spanning \textbf{500k} expert-involved multiple-choice questions and \textbf{50k} regression tasks, the benchmark enables comprehensive assessment of both decision-level and estimation-level bias in Vision--Language Models (VLMs). Evaluations on SDGBiasBench reveal an intrinsic \textbf{SDG bias} in current VLMs, where predictions are frequently driven by SDG-specific priors rather than reliable multi-modal cues. To mitigate such bias, we propose \textbf{CADE}~(\textbf{C}ontrastive \textbf{A}daptive \textbf{D}ebias \textbf{E}nsemble), a training-free, plug-and-play method that leverages modality-specific answer priors. CADE yields significant gains on the proposed benchmark, improving multiple-choice accuracy by up to 25\% and reducing regression MAE by up to 12 points across multiple VLMs. We hope our work can foster the development of more fair and reliable AI systems for sustainable development.\ifreview\else\ Code is available at: \href{https://github.com/jeffreylin122/SDGBiasBench}{SDGBiasBench}.\fi

\end{abstract}

\section{Introduction}
\label{sec:introduction}

Established by the United Nations, Sustainable Development Goals (SDGs) define a shared global agenda for sustainable and inclusive development across social, economic, and environmental dimensions \citep{un2030agenda2015}. Vision--language foundation models \citep{bommasani2021foundation,radford2021learning,alayrac2022flamingo,li2023blip,dai2023instructblip,liu2023llava,liu2024llava15,chen2023palix,bai2025qwen25vl,chen2024internvl25,li2024llavaonevision} provide a promising paradigm for SDG-related monitoring by jointly leveraging visual and textual signals, complementing prior sustainability work that uses satellite imagery to measure development outcomes \citep{jean2016combining,burke2021using,yeh2021sustainbench}.

Prior work has documented geographic and socio-economic biases in foundation models used in sustainability and geospatial settings \citep{manvi2024geobias,huang-etal-2025-ai,pouget2024nofilter}. Since SDG applications are closely related to global development, resource allocation, and policy-making, biased behavior in these models risks amplifying inequities and producing unreliable evidence \citep{bender2021stochastic,mitchell2019model,hardt2016equality}. Ensuring fair and trustworthy AI for sustainable development therefore necessitates a dedicated framework to systematically evaluate and address such biases \citep{bommasani2021foundation,mitchell2019model}.

To this end, we propose \textbf{SDGBiasBench}, a large-scale benchmark suite for vision–language reasoning over SDG-related indicators, explicitly designed to diagnose sustainability-oriented bias \citep{yeh2021sustainbench,manvi2024geobias}. The benchmark comprises expert-involved, multi-step, multi-modal questions spanning both qualitative judgments (\eg risk levels, comparative conditions) and quantitative estimation (\eg rates, indices, magnitudes), enabling comprehensive evaluation of model performance on sustainability-oriented prediction tasks \citep{antol2015vqa,hudson2019gqa,lu2022scienceqa,fu2023mme}. By enforcing multi-step, multi-modal reasoning and evaluating behavior under controlled evidence conditions \citep{niu2021counterfactual,zhang2022counterbias,zhang2025debiasing}, SDGBiasBench exposes whether predictions track the provided inputs or remain dominated by bias.

We empirically reveal a pronounced \textbf{SDG bias} in Vision--Language Models (VLMs) that manifests as pillar-conditioned directional defaults and imbalanced multi-modal grounding \citep{manvi2024geobias,zhou2022vlstereoset,ruggeri2023multidimensional,weng2024imagesspeak} using SDGBiasBench. Across the three pillars in SDGBiasBench, VLMs exhibit systematic optimistic, conservative, or pessimistic leanings that are triggered differently by health, infrastructure, and human-capital indicators, forming distinctive bias signatures \citep{zhou2022vlstereoset,ruggeri2023multidimensional}. Meanwhile, even when imagery contains informative spatial cues (\eg settlement patterns, road connectivity, land use), models often under-weight visual evidence relative to textual cues \citep{zhang2025debiasing,leng2024vcd,li2023pope}, leading predictions to follow entrenched geographic correlations rather than grounded inspection \citep{manvi2024geobias,burke2021using}.

To mitigate such bias, we introduce a simple yet effective training-free, plug-and-play approach \textbf{CADE}~(\textbf{C}ontrastive \textbf{A}daptive \textbf{D}ebias \textbf{E}nsemble),
which detects over-reliance on priors by comparing predictions from different constrained input views, adaptively reweighting outputs to favor visual evidence over context \citep{niu2021counterfactual,zhang2025debiasing,leng2024vcd,janghorbani2023multi,weng2024images}. Without introducing training overhead during model inference, CADE yields remarkable gains, improving multiple-choice accuracy and reducing regression MAE significantly across multiple VLMs, promoting more fair and evidence-based predictions.

In summary, our contributions are:
\vspace{-3pt}
\begin{enumerate}
    \vspace{-3pt}
    \item \textbf{Benchmark Suite for Revealing SDG Bias}:
    We introduce \textbf{SDGBiasBench}, the first large-scale benchmark suite for both qualitative judgments and quantitative estimation on diagnosing bias across Sustainable Development Goal (SDG) indicators.

    \vspace{-3pt}
    \item \textbf{Identification of SDG Bias}:
    Our analysis reveals pillar-conditioned directional defaults and modality imbalance in SDG reasoning, raising fairness concerns for SDG applications.

    \vspace{-3pt}
    \item \textbf{Training-free Debiasing Baseline}:
    We propose \textbf{CADE}, a plug-and-play debiasing method that adaptively reweights multiple input views to mitigate SDG bias thus improve performance, serving as a simple-but-effective baseline.

\end{enumerate}
\section{SDGBiasBench: An Overview}
\label{sdgbiasbench}

\begin{figure*}[t]
    \centering
    \includegraphics[width=0.96\textwidth]{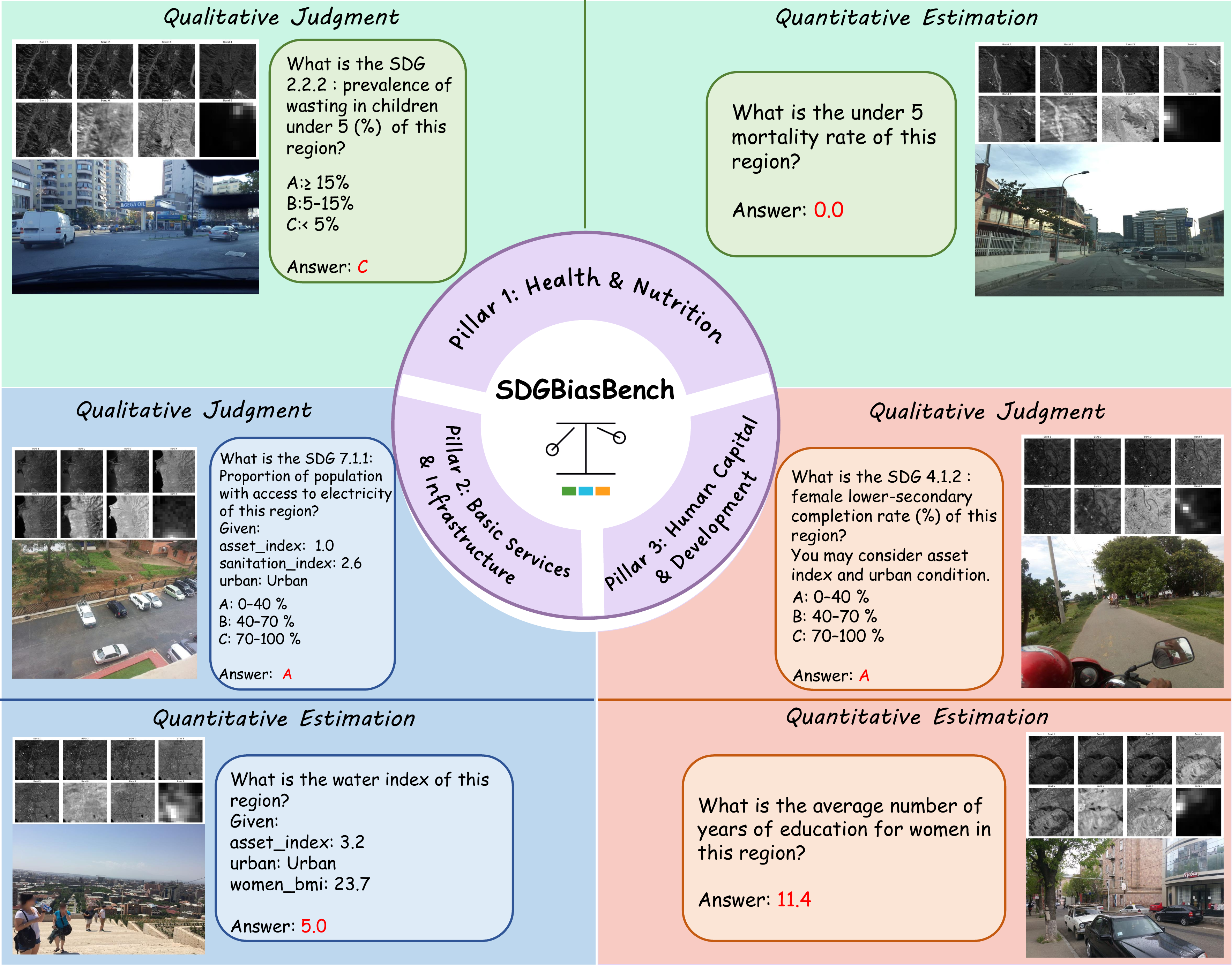}
    \vspace{-3pt}
    \caption{\textbf{Overview of \textit{SDGBiasBench}}. The three sustainability pillars are each illustrated with one qualitative judgment and one quantitative estimation example, showcasing the multi-modal SDG reasoning tasks used to probe SDG biases in VLMs.}
    \label{fig:sdgbias_example}
    \vspace{-15pt}
\end{figure*}

SDGBiasBench aims to evaluate vision–language reasoning on real-world Sustainable Development Goals (SDGs) related indicators to reveal SDG biases. As illustrated in~\Cref{fig:sdgbias_example}, we organize the benchmark into three pillars: \textbf{\emph{Health \& Nutrition}}, \textbf{\emph{Basic Services \& Infrastructure}}, and \textbf{\emph{Human Capital \& Development}}, each grouping indicators with shared real-world measurement and evidence patterns. The benchmark comprises two task types: qualitative judgments and quantitative estimation to mirror real-world scenarios. Specifically, we use multiple-choice questions for qualitative judgments and regression tasks for quantitative estimation, respectively. Each task is paired with satellite imagery, structured context variables (\eg sanitation index, women’s education level, asset index) and often augmented with street-level photos, requiring multi-step reasoning across modalities.

The three pillars capture distinct but complementary forms of sustainable development reasoning.
\textbf{\emph{Health \& Nutrition}} focuses on inferring latent health conditions, nutritional status, and healthcare access from indirect environmental and socio-economic cues. \textbf{\emph{Basic Services \& Infrastructure}} centers on access to essential services such as electricity, water, and sanitation, requiring models to relate infrastructure and environmental patterns to underlying service quality and deprivation. \textbf{\emph{Human Capital \& Development}} targets broader reasoning on education and composite development outcomes, where signals are more diffuse and must be inferred from holistic socio-economic context.

Human experts are involved to provide high-quality answers for diverse reasoning tasks for sustainable development that require grounded multi-modal inference beyond surface-level perception.  
We provide more details in~\Cref{sec:appendix_sdgbiasbench} including annotation protocols, dataset statistics, and benchmark composition and quality-control details.

\vspace{-6pt}

\section{Empirical Analysis of SDG Bias}
\label{sec:sdg_bias_analysis}

\vspace{-6pt}
In this section, we use SDGBiasBench as a diagnostic lens to evaluate VLMs' behaviors on SDG-related questions. Although reliance on language priors is a known phenomenon in VLMs, our findings suggest an SDG-specific mechanism, which we define as \textbf{SDG Bias}: \emph{when inferring the SDG indicators, models (i) produce \textbf{pillar-conditioned directional defaults}: systematically being optimistic/conservative/pessimistic in ways that differ across health, infrastructure, and human capital; (ii) exhibit \textbf{modality-imbalanced evidence use}: treating structured SDG context as largely sufficient and under-utilizing aligned imagery.} These signatures motivate our view-aware debiasing method, which will be further elaborated in~\Cref{sec:cade}.

\vspace{-3pt}
\subsection{Setup: Views, Models, and Metrics}
\label{subsec:analysis_setup}

We evaluate three representative 7B-parameter VLMs: \textsc{LLaVA-v1.5}~\citep{liu2024llava15}, \textsc{InstructBLIP}~\citep{dai2023instructblip}, \textsc{Qwen2.5-VL}~\citep{bai2025qwen25vl} on SDGBiasBench under four \emph{input views}. For each datapoint $i$, let $I_i$ denote the image(s) (satellite and/or street-level), $C_i$ the structured context (auxiliary variables/indices), and $Q_i$ the question. We define the four inference prompts:

\begin{itemize}
    \vspace{-3pt}
    \item \textbf{Q-only} ($T^{\text{q}}$): $(Q_i)$
    \vspace{-3pt}
    \item \textbf{CTX+Q} ($T^{\text{ctx}}$): $(C_i, Q_i)$
    \vspace{-3pt}
    \item \textbf{IMG+Q} ($T^{\text{img}}$): $(I_i, Q_i)$
    \vspace{-3pt}
    \item \textbf{Full} ($T^{\text{full}}$): $(I_i, C_i, Q_i)$
\end{itemize}

\begin{figure}[t]
    \centering
    \includegraphics[width=0.99\linewidth]{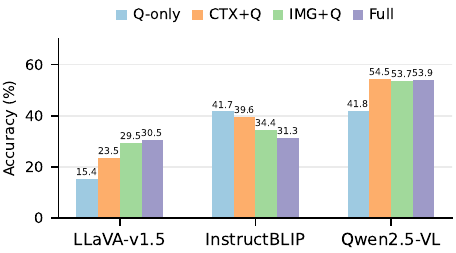}
    \vspace{-15pt}
    \caption{\textbf{Per-view MCQ Accuracy.} Accuracy (\%) for three VLMs under four evidence views (Q-only, CTX+Q, IMG+Q, Full).}
    \vspace{-6pt}
    \label{fig:sdgbias_per_view_acc}
\end{figure}

\vspace{-6pt}
\noindent In the main text, we use the multiple-choice questions (MCQs) as our primary diagnostic signal and report \textbf{Accuracy} as the metric.\footnote{More experimental details are provided in~\cref{sec:t1_t5_explain}.} 
A summary of per-view MCQ performance is given in~\Cref{fig:sdgbias_per_view_acc}. 

\vspace{-3pt}
\subsection{Pillar-conditioned Directional Defaults Form an SDG Bias Signature}
\label{subsec:finding_directional_pillar}

SDG bias manifests as \textbf{directional defaults} that are \textbf{pillar-conditioned}.  To isolate \textbf{directional SDG bias} from evidence use, we examine the \textbf{Q-only} setting, where an evidence-grounded model should have no systematic reason to prefer any particular answer direction (\ie roughly a 33\%-each distribution for a three-option question). In our MCQ tasks, the candidate options are \emph{ordinal}, allowing us to map outputs into three directional categories: \textbf{Pessimistic} (lower-development choice), \textbf{Conservative} (middle choice), and \textbf{Optimistic} (higher-development choice). Accordingly, these directional categories do not measure correctness in Q-only, but rather diagnose the direction in which the model tends to guess when no supporting evidence is available.

\begin{figure}[t]
    \centering
    \includegraphics[width=0.99\linewidth]{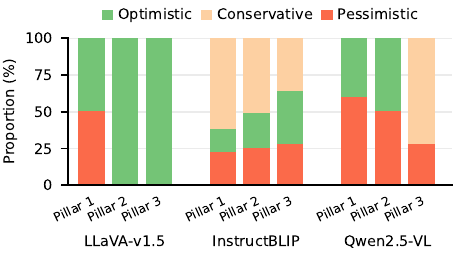}
    \vspace{-15pt}
    \caption{\textbf{Distribution of Q-only prediction tendencies across pillars.} Results reveal directional defaults without visual evidence.}
    \vspace{-15pt}
    \label{fig:sdgbias_outcome_qonly}
\end{figure}

Figure~\ref{fig:sdgbias_outcome_qonly} shows that these priors are not only model-specific but also \emph{pillar-dependent}, forming distinctive bias signatures across Health \& Nutrition, Basic Services \& Infrastructure, and Human Capital \& Development. Concretely, each model exhibits a characteristic triplet of outcome distributions across the three pillars, revealing where it systematically leans pessimistic, anchors to the middle, or defaults optimistic.
We discuss each VLM's preference in detail as follows:

\paragraph{\textsc{LLaVA-v1.5}: Polarized Health, Near-Deterministic Optimism Elsewhere.}
\textsc{LLaVA-v1.5-7B} displays a strongly pillar-conditioned pattern with minimal conservative anchoring. On Pillar~1, it is bimodal: responses split roughly evenly between pessimistic and optimistic options. In contrast, on Pillar~2 and Pillar~3, it becomes overwhelmingly \textbf{optimistic}, suggesting that for infrastructure and human-capital questions, the model carries a strong upward default even without supporting evidence.

\paragraph{\textsc{InstructBLIP}: Conservative Anchoring with Moderated Asymmetry.}
\textsc{InstructBLIP-7B} exhibits a comparatively stable conservative default across pillars, consistently placing the largest mass on the middle option. This conservative anchoring is the strongest on Pillar~1 (majority conservative, with smaller pessimistic and optimistic fractions), remains dominant on Pillar~2, and becomes more mixed on Pillar~3 where the distribution is closer to a three-way split. Overall, \textsc{InstructBLIP-7B} behaves like a "centrist" prior under missing evidence: it tends to hedge toward the midpoint, while still allowing pillar-dependent skews in the tails.

\paragraph{\textsc{Qwen2.5-VL}: Pessimistic on Health and Infrastructure, Conservative on Human Capital.}
\textsc{Qwen2.5-VL-7B} shows a pronounced pessimistic leaning on Pillar~1 and Pillar~2: for both pillars, the pessimistic category constitutes a large portion of outputs (roughly half or more), with the remainder largely optimistic and little reliance on the conservative option. The pattern changes sharply on Pillar~3, where the model shifts to a mostly \textbf{conservative} default and optimism nearly disappears. This pillar-sensitive switch indicates that the model activates different directional priors depending on the SDG theme, rather than applying a uniform "cautious" strategy.

\subsection{Modality Imbalance in SDG Reasoning}
\label{subsec:finding_modality_imbalance}

\begin{figure}[t]
    \centering
    \includegraphics[width=0.99\linewidth]{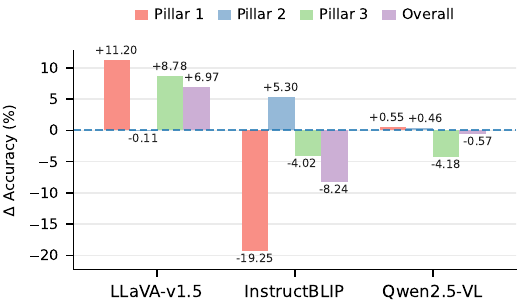}
    \vspace{-21pt}
    \caption{\textbf{Performance difference under different input views.} 
    Results of $\Delta$Acc\ (Full $-$ CTX+Q) show consistent modality imbalance. 
    }
    \vspace{-12pt}
    \label{fig:sdgbias_img_contrib}
\end{figure}

Furthermore, we observe a consistent \emph{modality imbalance} across all three pillars. 
As shown in~\Cref{fig:sdgbias_img_contrib}, once structured SDG context is provided, models often commit early, and adding aligned imagery yields only marginal gains (sometimes even degrading accuracy). This is notable because SDGBiasBench depends on \emph{latent} visual proxies: settlement morphology/density, road connectivity, and land-use/environmental cues, yet imagery is frequently under-weighted even when it should refine or correct the inference.
We discuss the observed imbalance in each pillar as follows:

\vspace{-3pt}
\paragraph{Pillar 1: Health \& Nutrition.}
On indicators like stunting prevalence (SDG~2.2.1), predictions largely stabilize under \textbf{CTX+Q}; imagery consistent with better living conditions (denser, more connected settlements, stronger access corridors) rarely overturns an activated "high-risk" interpretation.

\vspace{-3pt}
\paragraph{Pillar 2: Basic Services \& Infrastructure.}
For outcomes like electricity access (SDG~7.1.1), imagery should be highly diagnostic (built-up continuity, road hierarchy), but several VLMs show little change from \textbf{CTX+Q} to \textbf{Full}, indicating inference driven more by textual priors than spatial inspection.

\vspace{-3pt}
\paragraph{Pillar 3: Human Capital \& Development.}
For composites like Subnational Human Development Index (SHDI), models often default to the middle bin under context, with weak image-driven shifts despite diffuse proxies (urbanization gradients, corridor structure), suggesting context-conditioned priors dominate multi-modal integration.

\vspace{-6pt}
\section{CADE: Contrastive Adaptive Debiasing Ensemble}
\label{sec:cade}

To mitigate \textbf{SDG Bias}, we introduce \textbf{CADE}, a training-free, plug-and-play debiasing method that operates purely on
\emph{logits} from a VLM, without modifying the model parameters.\footnote{We provide further analysis and discussions on the uniqueness, significance, and superiority of CADE in~\cref{sec:additional_analysis}.}
We reuse the four
views $\{T^{\text{q}},T^{\text{ctx}},T^{\text{img}},T^{\text{full}}\}$ defined in
Section~\ref{subsec:analysis_setup}. For each datapoint $i$, we query the model under
these views and obtain per-option logits. We describe CADE for MCQ inference below.

\begin{figure}[t]
    \centering
    \includegraphics[width=0.99\columnwidth]{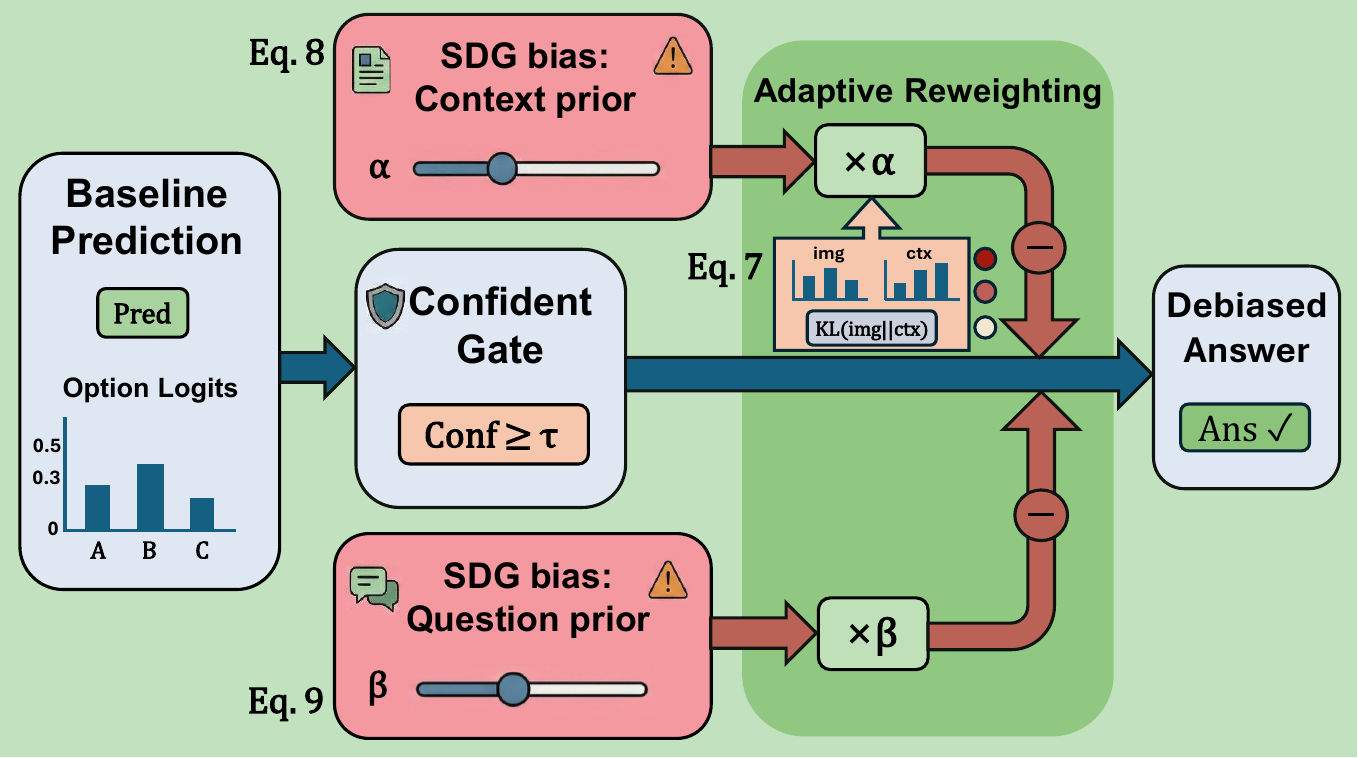}
    \vspace{-15pt}
    \caption{\textbf{Overview of \textit{CADE}}, which applies a confidence gate and, when triggered, adaptively reweights context and question SDG bias priors, scaling the context prior by the image context divergence, to produce a debiased answer.}
    \vspace{-12pt}
    \label{fig:cade}
\end{figure}

\paragraph{Candidates and logits.}
Let $\mathcal{O}_i$ denote the finite option set for datapoint $i$
(\eg $\{A,B\}$ or $\{A,B,C\}$). Under each view $T$, the VLM returns a pre-softmax
logit vector over $\mathcal{O}_i$:
\[
\mathbf{z}^{(T)}_i = \bigl(z^{(T)}_{i,a}\bigr)_{a \in \mathcal{O}_i}.
\]
Here we denote the collection of all view-wise logits for datapoint $i$ as
$\mathbf{Z}_i = \{\mathbf{z}^{(T)}_i\}_{T \in \{ \text{q},\text{ctx},\text{img},\text{full}\}}$.

\paragraph{Step 1: View-wise probabilities computation.}
We convert logits into probabilities with softmax:
\begin{equation}
  p^{(T)}_{i,a}
  =
  \frac{\exp\bigl(z^{(T)}_{i,a}\bigr)}
       {\sum_{a' \in \mathcal{O}_i}\exp\bigl(z^{(T)}_{i,a'}\bigr)},
  \qquad a \in \mathcal{O}_i.
  \label{eq:cade-softmax}
\end{equation}

\paragraph{Step 2: Confidence-gated thresholding.}
We treat the Full view $T^{\text{full}}$ as a conservative baseline:
\begin{equation}
  \hat{a}^{\text{base}}_i
  =
  \arg\max_{a \in \mathcal{O}_i}~p^{(\text{full})}_{i,a},
  \quad
  m_i
  =
  \max_{a \in \mathcal{O}_i}~p^{(\text{full})}_{i,a}.
  \label{eq:cade-base-conf}
\end{equation}
Given $\tau \in (0,1)$, we keep the baseline prediction when it is confident
(\ie $m_i \ge \tau$). Otherwise (when $m_i < \tau$), we invoke the debiasing ensemble below to produce the final answer.

\paragraph{Step 3: Three streams engagement (image / context / bias).}
We group the four views into three conceptual streams:
\begin{itemize}
  \item \emph{Image stream}: combines Full and IMG+Q,
  \item \emph{Context stream}: CTX+Q,
  \item \emph{Bias stream}: Q-only (intrinsic prior),
\end{itemize}
with the following calculations:
\begin{align}
  \tilde{p}^{\text{img}}_{i,a}
  &=
  p^{(\text{full})}_{i,a}
  +
  p^{(\text{img})}_{i,a},
  \label{eq:cade-pimg-tilde} \\
  \tilde{p}^{\text{ctx}}_{i,a}
  &=
  p^{(\text{ctx})}_{i,a},
  \label{eq:cade-pctx-tilde} \\
  \tilde{p}^{\text{q}}_{i,a}
  &=
  p^{(\text{q})}_{i,a}.
  \label{eq:cade-pq-tilde}
\end{align}
Next, we renormalize each stream to obtain the normalized probabilities:
\begin{equation}
  p^{s}_{i,a}
  =
  \frac{\tilde{p}^{s}_{i,a}}{\sum_{a' \in \mathcal{O}_i} \tilde{p}^{s}_{i,a'}},
  \quad
  s \in \{\text{img},\text{ctx},\text{q}\}.
  \label{eq:cade-stream-renorm}
\end{equation}

\paragraph{Step 4: Contrastive scoring with adaptive disagreement.}
We quantify image-context disagreement using KL divergence:
\begin{equation}
  D_i
  =
  \mathrm{KL}\bigl(p^{\text{img}}_{i}\,\Vert\,p^{\text{ctx}}_{i}\bigr)
  =
  \sum_{a \in \mathcal{O}_i}
  p^{\text{img}}_{i,a}
  \log
  \frac{p^{\text{img}}_{i,a}}{p^{\text{ctx}}_{i,a}}.
  \label{eq:cade-kl}
\end{equation}
For each datapoint $i$, we define an instance-wise coefficient $\alpha_i$ that controls the strength of the context penalty. It is determined by a base value $\alpha$ and a sensitivity factor $\lambda_{\mathrm{KL}}$, modulating the penalty according to the image-context KL divergence:
\begin{equation}
  \alpha_i
  =
  \alpha \bigl(1 + \lambda_{\mathrm{KL}} D_i \bigr).
  \label{eq:cade-alpha}
\end{equation}
Finally, we compute a CADE score for each candidate $a$ in $\mathcal{O}_i$ as:
\begin{equation}
{score}_{i,a}
=
\log p^{\text{img}}_{i,a}
-\alpha_i \log p^{\text{ctx}}_{i,a}
-\beta \log p^{\text{q}}_{i,a},
\ a \in \mathcal{O}_i.
\label{eq:cade-score}
\end{equation}
Here, $\beta \ge 0$ is introduced to control how strongly we penalize question-only priors. 
Therefore, the debiased prediction $\hat{a}^{CADE}_i$ is given as:
\begin{equation}
  \hat{a}^{CADE}_i = \arg\max_{a \in \mathcal{O}_i}~{score}_{i,a}.
  \label{eq:cade-argmax}
\end{equation}
The algorithm of this plug-and-play \textbf{CADE} is outlined in~\Cref{alg:cade}.

\begin{algorithm}[t]
\caption{CADE}
\label{alg:cade}
\begin{algorithmic}[1]
\REQUIRE Per-view logit vectors over candidates $\mathcal{O}_i$, obtained by querying the model under four input views $\{\mathbf{z}^{(\text{full})}_i,\mathbf{z}^{(\text{ctx})}_i,\mathbf{z}^{(\text{img})}_i,\mathbf{z}^{(\text{q})}_i\}$
\STATE Convert logits to probabilities $p^{(T)}_{i,\cdot}$ (Eq.~\ref{eq:cade-softmax})
\STATE Compute baseline confidence $m_i$ (Eq.~\ref{eq:cade-base-conf})
\STATE \textbf{if} $m_i \ge \tau$ \textbf{then return} $\hat{a}^{\text{base}}_i$
\STATE Form $p^{\text{img}}_{i,\cdot}, p^{\text{ctx}}_{i,\cdot}, p^{\text{q}}_{i,\cdot}$
(Eqs.~\ref{eq:cade-pimg-tilde}--\ref{eq:cade-stream-renorm})
\STATE Compute $D_i$ and $\alpha_i$
(Eqs.~\ref{eq:cade-kl}--\ref{eq:cade-alpha})
\STATE Compute ${score}_{i,a}$ and $\hat{a}^{\text{CADE}}_i$
(Eqs.~\ref{eq:cade-score}--\ref{eq:cade-argmax})
\ENSURE Final debiased prediction $\hat{a}^{\text{CADE}}_i$
\end{algorithmic}
\end{algorithm}

\paragraph{Hyperparameters.}
We tune $\tau$, $\lambda_{\mathrm{KL}}$, $\alpha$ and $\beta$ via random search on a held-out validation set. 

\paragraph{Implementation details.}
More details and the regression variant of CADE are further discussed in~\Cref{sec:appendix_cade_details}.

\begin{figure*}[t]
    \centering
    \begin{subfigure}[t]{\textwidth}
        \includegraphics[width=0.98\linewidth]{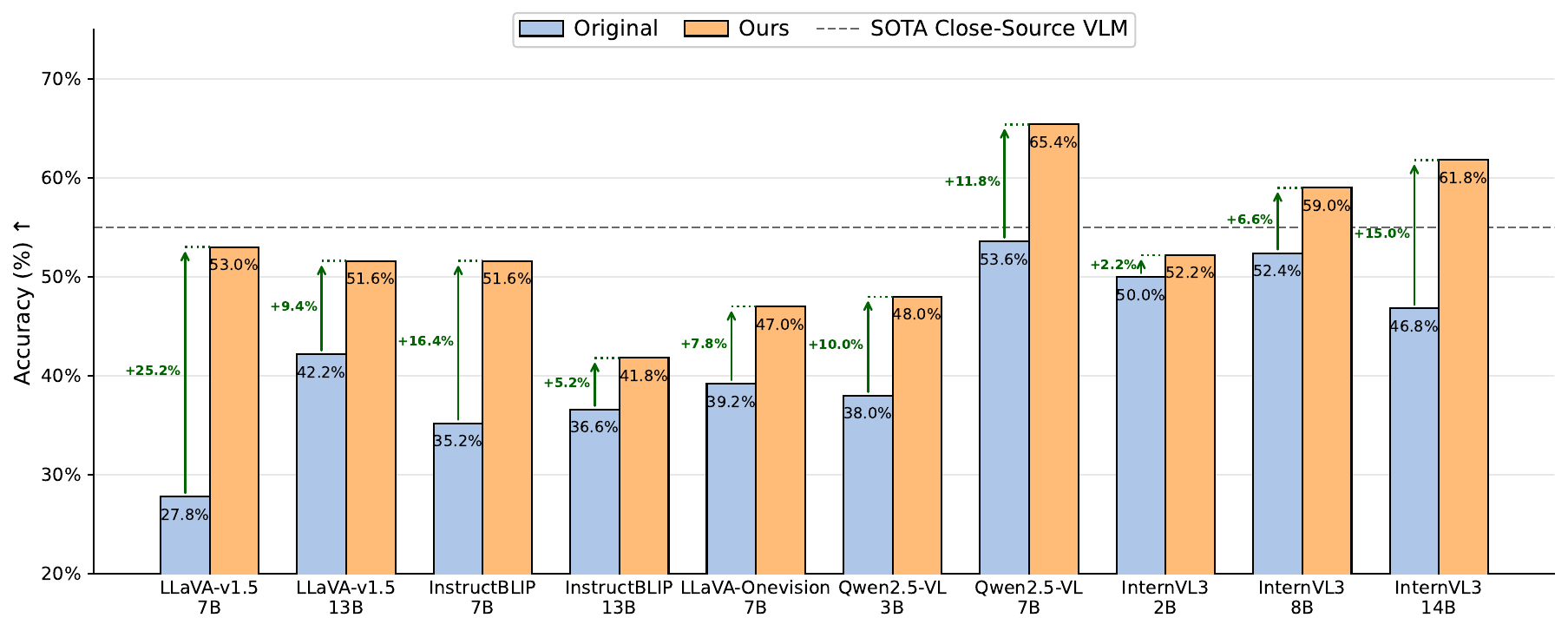}
        \label{fig:mcq_results_bar}
    \end{subfigure}
    \begin{subfigure}[t]{\textwidth}
        \includegraphics[width=0.98\linewidth]{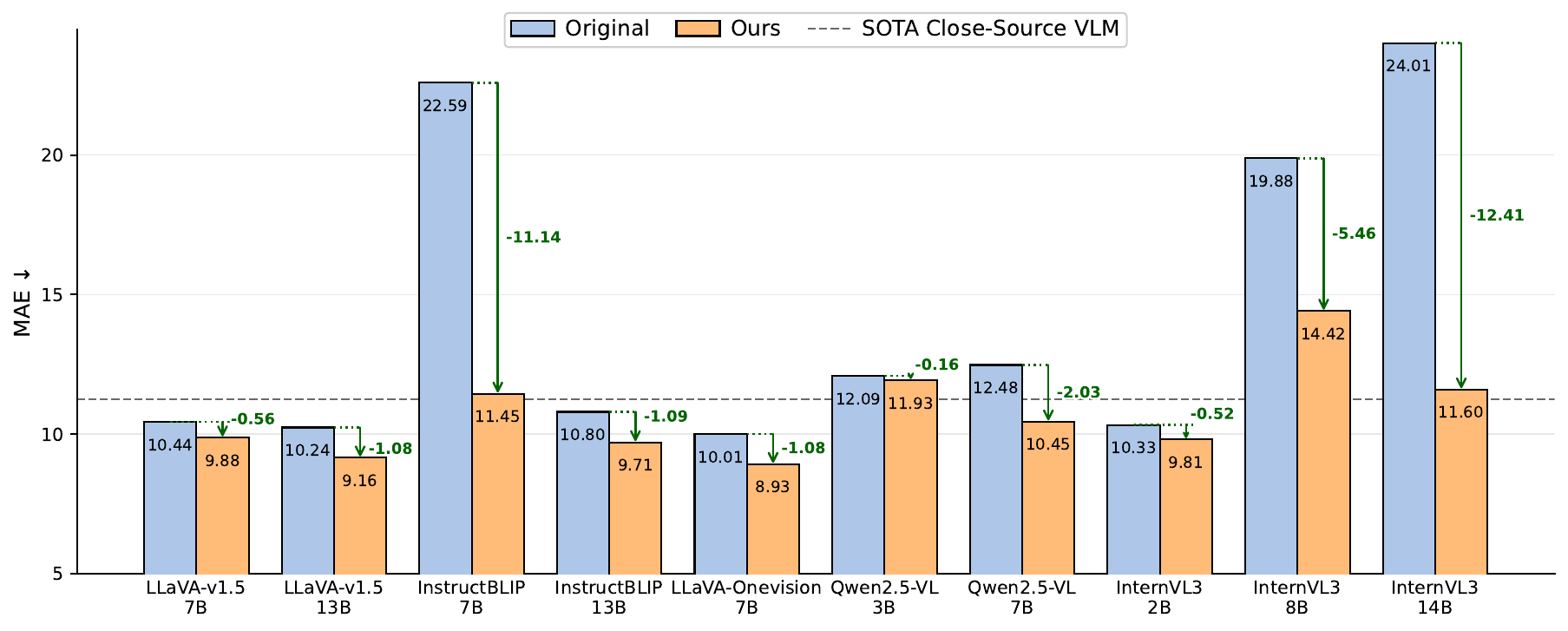}
        \label{fig:mae_results_bar}
    \end{subfigure}
    \vspace{-22pt}
\caption{\textbf{Performance of Vision--Language Models on: (Upper) multiple-choice questions; (Bottom) regression questions.} "Ours" refers to applying CADE to the same base model, where reductions are highlighted with green.}
\label{fig:mcq_mae_results_bar}
\vspace{-12pt}
\end{figure*}

\section{Experiments}
\label{sec:experiments}

This section serves two purposes: (i) to benchmark a diverse set of
vision--language models (VLMs) on \textbf{SDGBiasBench} under controlled
information regimes, and (ii) to evaluate \textbf{CADE} as a training-free plug-and-play method applied to each VLM.

\vspace{-4pt}
\subsection{Evaluation Protocol}
\label{subsec:exp_protocol}

\paragraph{Inputs and views.}
Each sample contains a satellite image, a street-level image (when available),
a structured text context, and a question.
We follow the four-view protocol defined in
Section~\ref{subsec:analysis_setup}: \textbf{Q-only}, \textbf{CTX+Q},
\textbf{IMG+Q}, and \textbf{Full}.\footnote{Unless otherwise stated, the "Original"~setting corresponds to standard inference with the \textbf{Full} input view.}

\vspace{-5pt}
\paragraph{Models.}
We evaluate representative open-source VLMs across various scales, which include: \textsc{LLaVA-v1.5}~\cite{liu2024llava15}, \textsc{InstructBLIP}~\cite{dai2023instructblip}, \textsc{LLaVA-Onevision}~\cite{li2024llavaonevision}, \textsc{Qwen2.5-VL}~\cite{bai2025qwen25vl}, and \textsc{InternVL3}~\cite{chen2025internvl3}.
Some well-known closed-source VLMs are also included for comprehensive comparison: \textsc{GPT-4o}~\cite{openai2024gpt4o}, \textsc{Gemini 2.0 Flash}~\cite{google2024gemini2}, and \textsc{Claude 3.5 Sonnet}~\cite{anthropic2024claude35}.

\vspace{-5pt}
\paragraph{Metrics.}
For \textbf{MCQ},  we report \textbf{Accuracy}.
For \textbf{Regression}, we report \textbf{MAE} (Mean Absolute Error) and \textbf{Interval Accuracy}. 
Interval Accuracy counts a correct prediction if it falls within a pre-defined tolerance interval for each question type.
We provide detailed interval definitions, per-interval results, and further in-depth analysis in~\Cref{sec:appendix_interval_accuracy}.

\begin{table}[t]
\centering
\caption{\textbf{Comparison with SOTA debiasing methods.}}
\vspace{-6pt}
\footnotesize
\renewcommand{\arraystretch}{1.12}
\setlength{\tabcolsep}{3pt}
\begin{tabular}{lcccc}
\toprule
\textbf{Model} & Original & Post-Hoc & VDD & \textbf{Ours} \\
\midrule
\textsc{LLaVA-v1.5-7B} & 0.28 & 0.34 & 0.41 & \textbf{0.53} \\
\textsc{Qwen2.5-VL-7B} & 0.54 & 0.57 & 0.58 & \textbf{0.65} \\
\bottomrule
\end{tabular}
\vspace{-12pt}
\label{tab:compare_with_other_debias_method}
\end{table}

\vspace{-5pt}
\subsection{Main Benchmark Results}
\label{subsec:exp_main_results}

\Cref{fig:mcq_mae_results_bar} shows MCQ accuracy and regression MAE results across VLMs. Across both tasks, the original performance is uniformly weak: MCQ accuracy remains low for most models, with several barely exceeding random-guess levels, while regression errors are consistently large, suggesting that current VLMs still struggle to make precise and dependable predictions on sustainability-oriented tasks.

\newcommand{\cmark}{\(\checkmark\)}
\newcommand{\xmark}{\(\text{--}\)}        

\begin{table}[t]
\centering
\caption{\textbf{Progressive ablation of CADE} on \textsc{LLaVA-v1.5-7B} (MCQ, T1). We progressively enable each CADE component and report accuracy.}
\vspace{-6pt}
\footnotesize
\renewcommand{\arraystretch}{1.12}
\setlength{\tabcolsep}{5pt}
\begin{tabular}{lcccc c}
\toprule
\textbf{Variant} & \multicolumn{4}{c}{\textbf{Hyperparameters}} & \textbf{Acc.} \\
\cmidrule(lr){2-5}
& $\alpha$ & $\lambda_{\mathrm{KL}}$ & $\beta$ & $\tau$ & \\
\midrule
\rowcolor{gray!15} Baseline & \xmark & \xmark & \xmark & \xmark & 0.30 \\
+ context penalty & \cmark & \xmark & \xmark & \xmark & 0.45 \\
+ adaptive disagreement & \cmark & \cmark & \xmark & \xmark & 0.46 \\
+ prior penalty & \cmark & \cmark & \cmark & \xmark & 0.52 \\
+ confidence gate (Ours) & \cmark & \cmark & \cmark & \cmark & \textbf{0.53} \\
\bottomrule
\end{tabular}
\vspace{-12pt}
\label{tab:cade_ablation}
\end{table}

\vspace{-3pt}
\subsection{CADE Results}
\label{subsec:exp_cade_results}

We apply CADE to open-source VLMs of different scales and report the same metrics as in the main results in~\Cref{fig:mcq_mae_results_bar}. 
Results\footnote{Detailed results and further discussion are in~\Cref{sec:appendix_regression_results}.} show that CADE consistently benefits all evaluated models, which raises MCQ accuracy across the board, and simultaneously lowers regression MAE for every model. Overall, the gains are substantial and stable across architectures and parameter scales, demonstrating that CADE is an effective and broadly applicable debiasing strategy.

\begin{figure*}[t]
\centering

{\small\textbf{\textsc{LLaVA-v1.5-7B} (MCQ, T1)}}\par\vspace{1mm}

\begin{minipage}{0.24\linewidth}
  \centering
  \includegraphics[width=\linewidth]{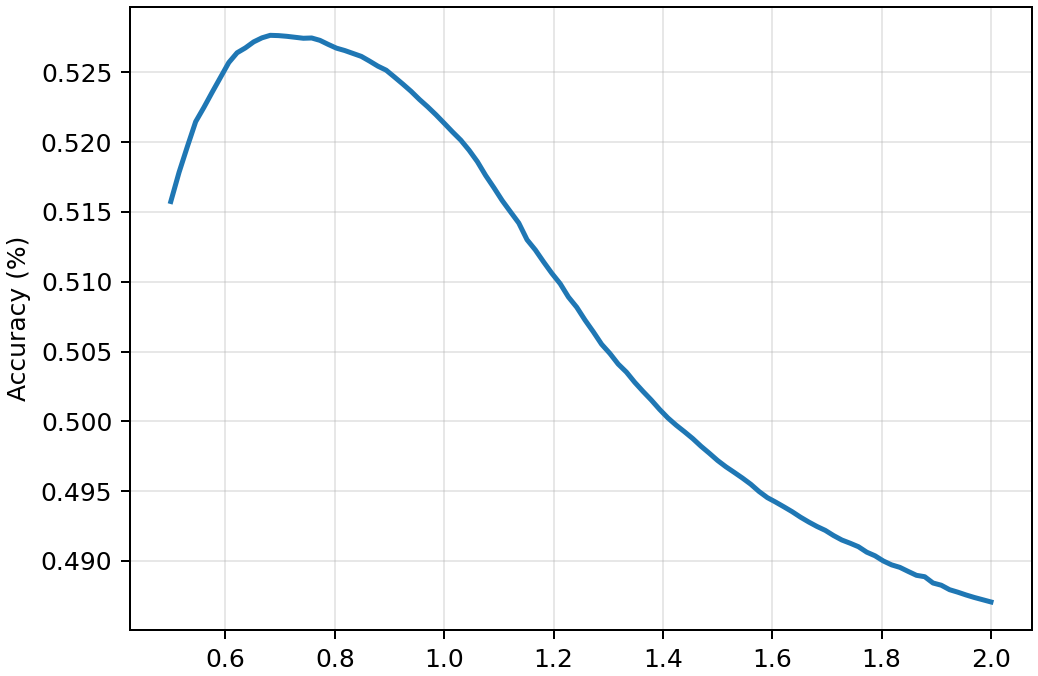}
  \vspace{-8mm}
  \caption*{\small (a) Sweep $\alpha$}
\end{minipage}\hfill
\begin{minipage}{0.24\linewidth}
  \centering
  \includegraphics[width=\linewidth]{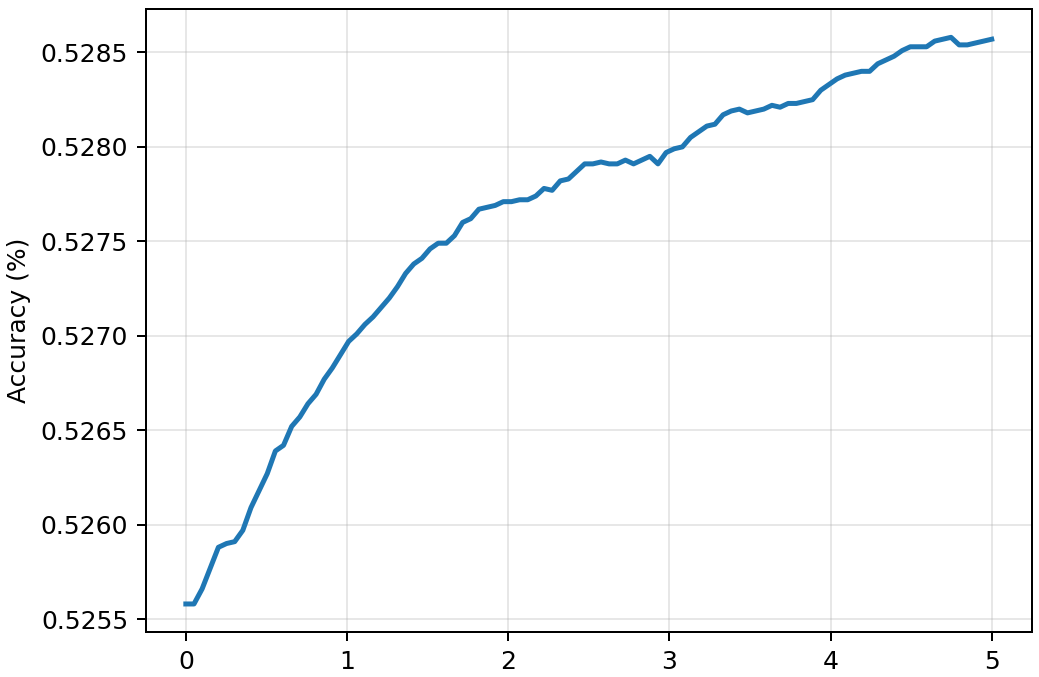}
  \vspace{-8mm}
  \caption*{\small (b) Sweep $\lambda_{\mathrm{KL}}$}
\end{minipage}\hfill
\begin{minipage}{0.24\linewidth}
  \centering
  \includegraphics[width=\linewidth]{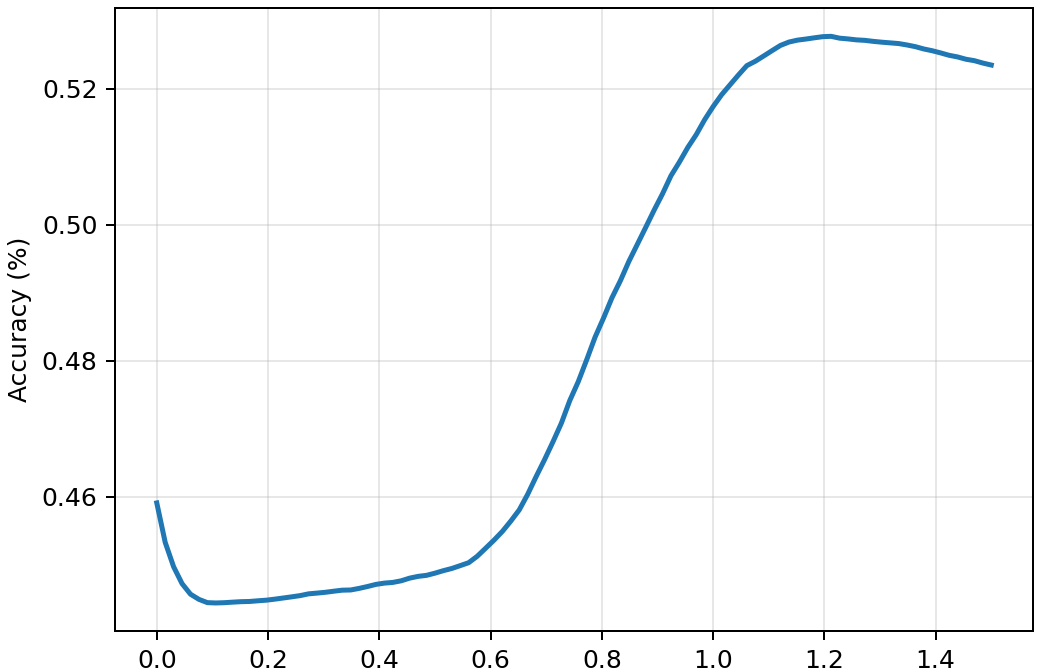}
  \vspace{-8mm}
  \caption*{\small (c) Sweep $\beta$}
\end{minipage}\hfill
\begin{minipage}{0.24\linewidth}
  \centering
  \includegraphics[width=\linewidth]{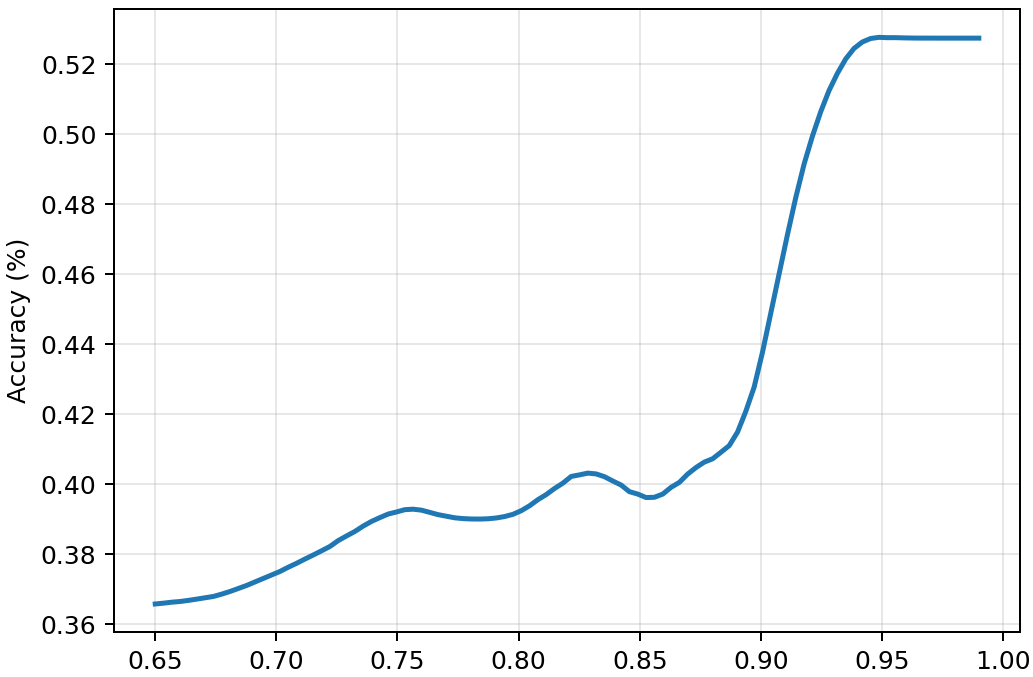}
  \vspace{-8mm}
  \caption*{\small (d) Sweep $\tau$}
\end{minipage}

\vspace{1mm}

{\small\textbf{\textsc{InstructBLIP-7B} (MCQ, T3)}}\par\vspace{1mm}

\begin{minipage}{0.24\linewidth}
  \centering
  \includegraphics[width=\linewidth]{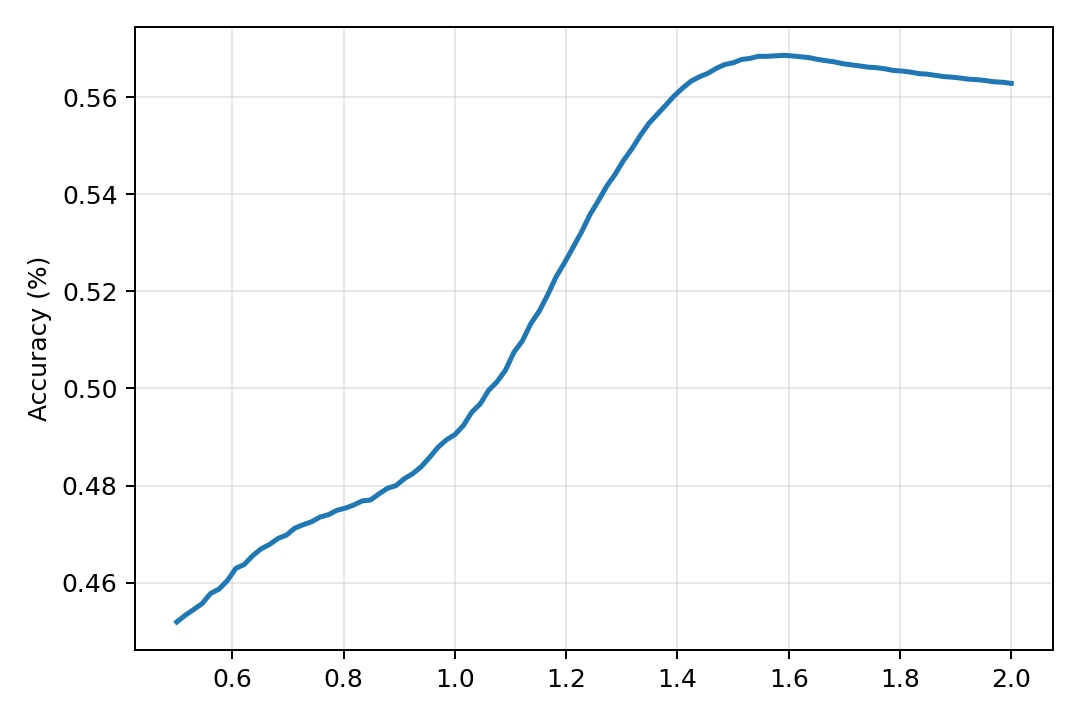}
  \vspace{-8mm}
  \caption*{\small (a) Sweep $\alpha$}
\end{minipage}\hfill
\begin{minipage}{0.24\linewidth}
  \centering
  \includegraphics[width=\linewidth]{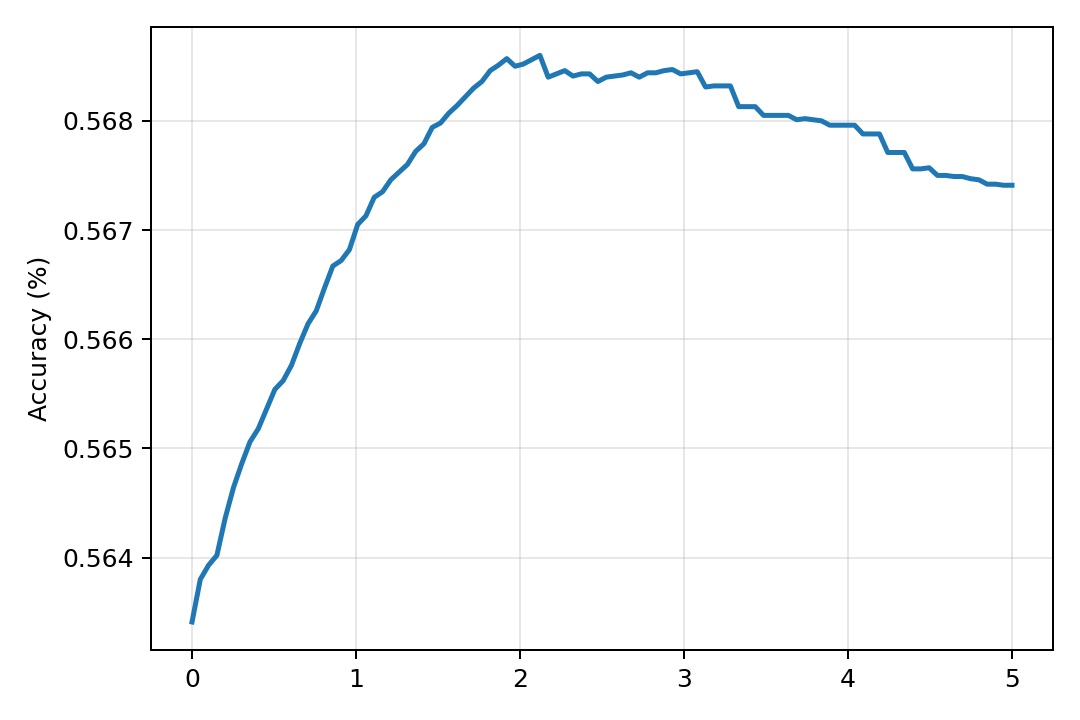}
  \vspace{-8mm}
  \caption*{\small (b) Sweep $\lambda_{\mathrm{KL}}$}
\end{minipage}\hfill
\begin{minipage}{0.24\linewidth}
  \centering
  \includegraphics[width=\linewidth]{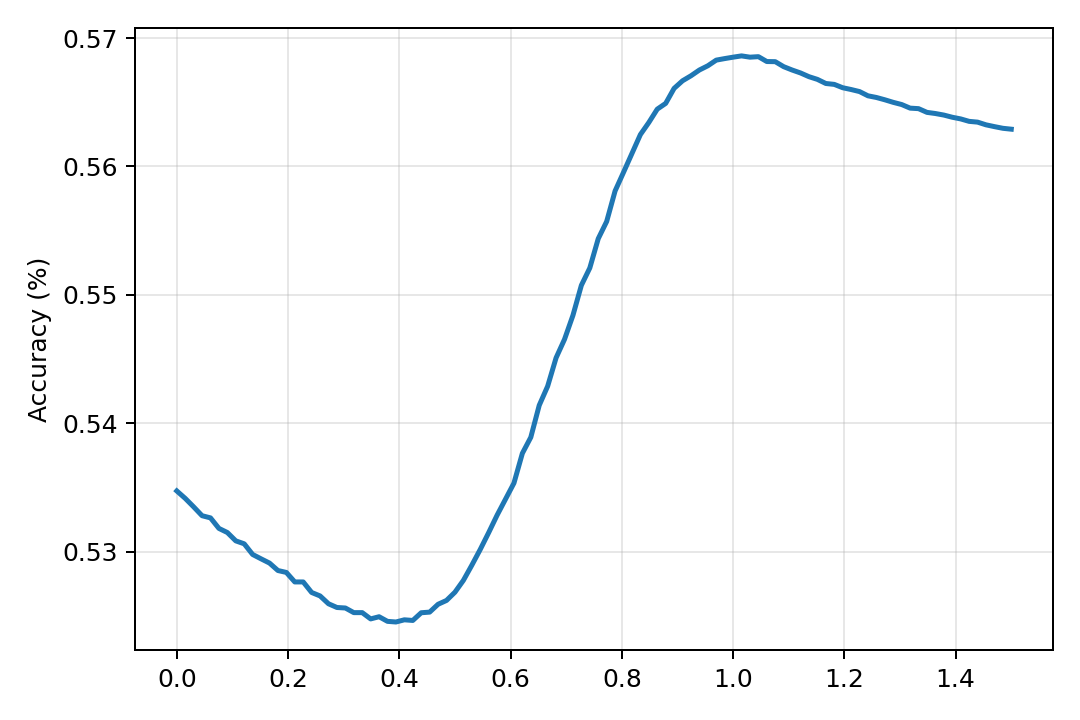}
  \vspace{-8mm}
  \caption*{\small (c) Sweep $\beta$}
\end{minipage}\hfill
\begin{minipage}{0.24\linewidth}
  \centering
  \includegraphics[width=\linewidth]{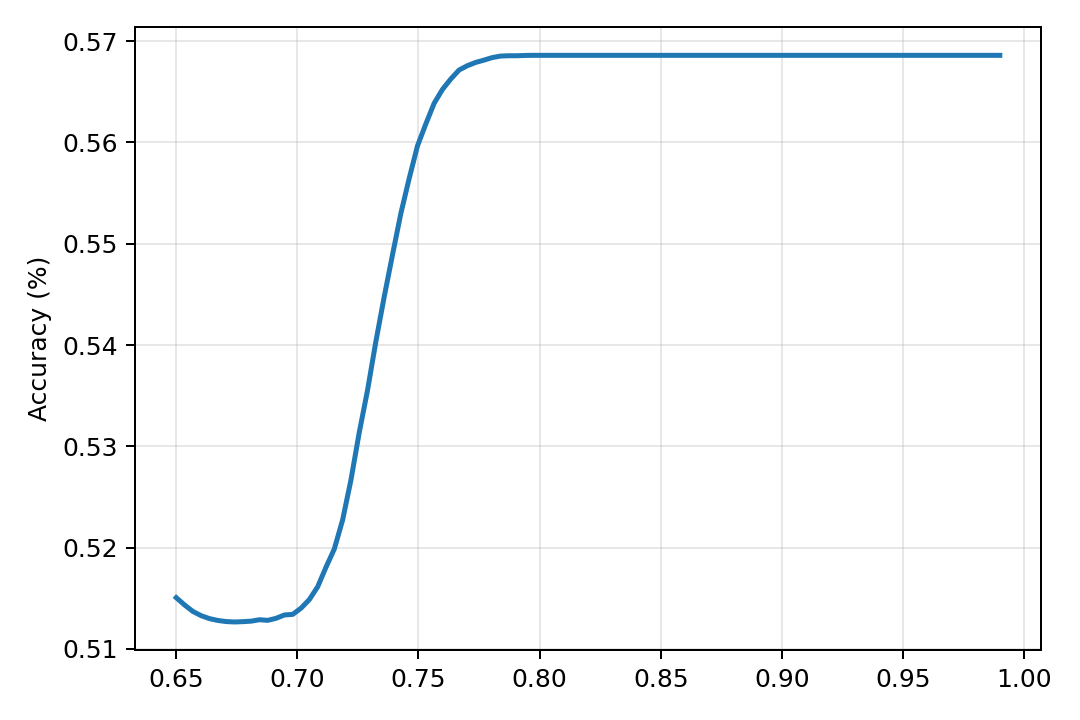}
  \vspace{-8mm}
  \caption*{\small (d) Sweep $\tau$}
\end{minipage}
\vspace{-5pt}
\caption{\textbf{Hyperparameter sensitivity study on \textsc{LLaVA-v1.5} and \textsc{InstructBLIP}.}
Each subplot varies one specific hyperparameter while fixing the other three.
MCQ accuracy is reported as VLM's performance.
}
\vspace{-3pt}
\label{fig:cade_sensitivity}
\end{figure*}

\begin{table*}[t]
\centering
\caption{\textbf{Debiasing component selection of CADE.} 
We compare different image-stream constructions under JS and KL divergence. 
``Full+Img.'' denotes the original CADE image stream obtained by normalizing \(p^{(\text{full})}+p^{(\text{img})}\).}
\vspace{-6pt}
\footnotesize
\renewcommand{\arraystretch}{1.12}
\setlength{\tabcolsep}{8pt}
\label{tab:cade-component-selection}
\begin{tabular}{lccccccc}
\toprule
\multirow{2}{*}{\textbf{Model}} 
& \multirow{2}{*}{\textbf{Original}} 
& \multicolumn{3}{c}{\textbf{JS}} 
& \multicolumn{3}{c}{\textbf{KL}} \\
\cmidrule(lr){3-5} \cmidrule(lr){6-8}
& 
& \textbf{Img.} 
& \textbf{Full} 
& \textbf{Full+Img.} 
& \textbf{Img.} 
& \textbf{Full} 
& \textbf{Full+Img.} \\
\midrule
\textsc{LLaVA-v1.5-7B} 
& 0.28 
& 0.49 & 0.48 & 0.50 
& 0.51 & 0.50 & \textbf{0.53} \\

\textsc{Qwen2.5-VL-7B} 
& 0.54 
& 0.59 & 0.61 & 0.63 
& 0.58 & 0.60 & \textbf{0.65} \\
\bottomrule
\end{tabular}
\vspace{-9pt}
\end{table*}

\vspace{-6pt}
\subsection{Comparison with SOTA Methods}
\label{subsec:exp_sota_compare}
We additionally provide comparisons against SOTA general debiasing methods, including Post-Hoc and VDD \citep{zhang2025debiasing} in~\Cref{tab:compare_with_other_debias_method}. Across models and evaluation settings, CADE consistently delivers the largest and most reliable improvements, showing that it is not only competitive with general-purpose debiasing baselines but also particularly effective at mitigating SDG-specific biases.

\vspace{-3pt}
\subsection{Ablation Studies}
\label{subsec:exp_ablation_sensitivity}

We conduct the following ablation studies\footnote{All ablations below are conducted with MCQ on one context test. Further details on these are given in~\cref{sec:t1_t5_explain}.} regarding the proposed CADE.
Further in-depth analysis and discussions on performance gain and the method's efficiency are provided in~\Cref{sec:additional_analysis}.

\vspace{-3pt}
\paragraph{Component analysis.} 
We provide a progressive ablation of CADE on \textsc{LLaVA-v1.5-7B} (MCQ, T1) in~\Cref{tab:cade_ablation}.
Results highlight the contribution of each step with consistent gains when components are gradually enabled:
adding the context penalty yields the largest performance increase (0.30$\rightarrow$0.45),
adaptive disagreement provides a small additional boost of +0.01, 
prior penalty brings a further improvement of +0.06, 
and the confidence gate gives a final increment of +0.01.

\vspace{-3pt}
\paragraph{Sensitivity study.} 
We conduct a sensitivity study of CADE as shown in~\Cref{fig:cade_sensitivity}.
Results show CADE is not brittle: accuracy peaks at moderate $\alpha$ ($\sim$0.7 for \textsc{LLaVA}, higher for \textsc{InstructBLIP}), rises with $\lambda_{\mathrm{KL}}$ with diminishing returns, and favors larger $\beta$ (best $\sim$1.0-1.2). $\tau$ instead has a sharp threshold ($\sim$0.9 \vs $\sim$0.75) then plateaus.

\vspace{-3pt}
\paragraph{Debiasing component selection.} 
\label{sec:component_selection}
We ablate two CADE design choices: the image stream \(p^{\text{img}}_i\) and the divergence in Eq.~\ref{eq:cade-kl}. For the image stream, we compare Image, Full, and our default Full+Image reference 
\((\tilde{p}^{\text{img}}_{i,a}=p^{(\text{full})}_{i,a}+p^{(\text{img})}_{i,a})\), followed by normalization.
We also replace KL with Jensen--Shannon (JS) divergence.
As shown in \Cref{tab:cade-component-selection}, Full+Image performs best under both JS and KL, suggesting that it gives a more reliable image-present reference than either single-view variant.
Since KL is slightly better and more consistent overall, we adopt Full+Image with KL as the default CADE design.

\section{Related Work}
\label{sec:related}

\paragraph{Geographic benchmarks for SDG monitoring.}
Earth observation has long supported large-scale estimation of development outcomes from satellite imagery \citep{henderson2012measuring,jean2016combining,head2017human,yeh2020publicly,burke2021using}. \textsc{SustainBench} systematizes SDG-relevant tasks and evaluation protocols \citep{yeh2021sustainbench}, and subsequent benchmark suites broaden coverage across sensors, tasks, and geographic regions \citep{marsocci2024pangaea,bountos2023fomo}. Remote-sensing vision--language benchmarks further study geospatial perception and grounded dialogue \citep{kuckreja2024geochat,danish2025geobenchvlm,soni2025earthdial}. However, prior geographic benchmarks rarely support both realistic SDG workflows and bias diagnosis on modern VLMs; \emph{SDGBiasBench} addresses these gaps by controlling evidence availability (Q/CTX/IMG/Full) for SDG-indicator reasoning on both qualitative judgments and quantitative estimation.

\vspace{-3pt}
\paragraph{Social and geographic bias in VLMs.}
Geographic stereotypes are documented even in text-only LLMs \citep{manvi2024geobias}, and multi-modal models inherit related skews from data filtering and representation gaps \citep{pouget2024nofilter}. For image geolocation, \citet{huang-etal-2025-ai} show systematic performance disparities favoring wealthy regions. Complementary analyses measure demographic and social biases in vision--language representations and systems \citep{ross2021groundedbias,ruggeri2023multidimensional,huang2025visbias}, building on bias diagnostics developed in NLP \citep{blodgett2020critical,nangia2020crows,nadeem2021stereoset}. Our work operationalizes these concerns in a sustainability setting as SDG-specific, pillar-conditioned \emph{directional defaults}.

\vspace{-3pt}
\paragraph{Training-free debiasing for VLMs.}
VLM decisions can be dominated by language priors when visual and textual cues conflict \citep{lin2024languagepriors}. A line of training-free methods mitigates such failures by contrasting decoding distributions, including VCD~\citep{leng2024vcd}, instruction-contrastive decoding~\citep{wang2024icd,chuang2309decoding,li2023contrastive,zhu2025ibd}, and explicit penalization of language priors~\citep{zhang2025debiasing}. 
CADE follows this training-free framework but targets SDG bias: it detects prior reliance via multi-view disagreement and adaptively reweights logits to reduce systematic directional errors.

\raggedbottom
\section{Conclusion}
\label{sec:conclusion}

In this work, we introduce a large-scale \textbf{SDGBiasBench} benchmark covering both qualitative MCQ judgments and quantitative regression estimation for SDG-indicator reasoning. 
Extensive evaluations reveal an intrinsic \textbf{SDG bias} across VLMs, characterized by pillar-conditioned directional defaults and persistent prior-driven modality imbalance. 
To mitigate this, we proposed \textbf{CADE}, a plug-and-play debiasing method that reweights outputs from multiple input views. 
CADE achieves significant gains on the proposed benchmark, supporting more reliable sustainability-oriented predictions.

\section*{Limitations}
\label{sec:limitations}

This work focuses on three major pillars and a representative set of indicators, where some lack-of-definition and outdated SDG targets and domains are not covered. 
Moreover, a small fraction of examples may carry imperfect annotations due to upstream label noise inherited from \textsc{SustainBench}; we treat these as part of the realistic evaluation setting rather than artifacts introduced during the benchmark construction. 
Finally, the proposed CADE is in principle applicable to closed-source VLMs, but it relies on per-candidate logits (or equivalent scores), which are typically not available by current usage of APIs from closed-source VLMs, preventing a direct end-to-end evaluation.

\section*{Ethical Considerations}
\label{sec:ethical_considerations}

SDG indicators are high-stakes: biased predictions (\eg systematic over/under-estimation across regions) can lead to harmful downstream decisions if used for policy or aid allocation. SDGBiasBench is intended for \emph{evaluation and bias diagnosis} rather than direct deployment, and our design explicitly probes prior reliance and directional error under changing evidence. Following \textsc{SustainBench}~\citep{yeh2021sustainbench}, privacy risks are mitigated by the DHS cluster-level release protocol and randomized geolocation jittering; imagery sources are publicly available and largely low-resolution, while street-level photos (when present) are privacy-protected (\eg automated blurring of faces/plates). We therefore expect limited risk of identifying individuals from the released benchmark, but emphasize that users should follow the original data providers' terms and apply appropriate safeguards when building models.

\ifreview
\else
\section*{Acknowledgments}
This work was conducted in collaboration with ST Engineering Geo-Insights Pte Ltd. The authors sincerely thank the reviewers, area chairs, and program chairs for their valuable feedback and constructive comments. 
\vspace{-0.5\baselineskip}
\fi

{
\bibliography{sections/09_references}

@article{alayrac2022flamingo,
  title   = {Flamingo: a visual language model for few-shot learning},
  author  = {Alayrac, Jean-Baptiste and Donahue, Jeff and Luc, Pauline and Miech, Antoine and Barr, Iain and Hasson, Yana and Lenc, Karel and Mensch, Arthur and Millican, Katherine and Reynolds, Malcolm and others},
  journal = {Advances in neural information processing systems},
  volume  = {35},
  pages   = {23716--23736},
  year    = {2022}
}

@misc{anthropic2024claude35,
  title  = {Claude 3.5 Sonnet Model Card},
  author = {{Anthropic}},
  year   = {2024},
  url    = {https://assets.anthropic.com/m/61e7d27f8c8f5919/original/Claude_3_Model_Card.pdf}
}

@inproceedings{antol2015vqa,
  title     = {Vqa: Visual question answering},
  author    = {Antol, Stanislaw and Agrawal, Aishwarya and Lu, Jiasen and Mitchell, Margaret and Batra, Dhruv and Zitnick, C Lawrence and Parikh, Devi},
  booktitle = {Proceedings of the IEEE international conference on computer vision},
  pages     = {2425--2433},
  year      = {2015}
}

@inproceedings{atabuzzaman2025benchmarking,
  title     = {Benchmarking and Mitigating MCQA Selection Bias of Large Vision-Language Models},
  author    = {Atabuzzaman, Md and Asgarov, Ali and Thomas, Chris},
  booktitle = {Proceedings of the 2025 Conference on Empirical Methods in Natural Language Processing},
  pages     = {33536--33550},
  year      = {2025}
}

@article{bai2025qwen25vl,
  title   = {Qwen2. 5-vl technical report},
  author  = {Bai, Shuai and Chen, Keqin and Liu, Xuejing and Wang, Jialin and Ge, Wenbin and Song, Sibo and Dang, Kai and Wang, Peng and Wang, Shijie and Tang, Jun and others},
  journal = {arXiv preprint arXiv:2502.13923},
  year    = {2025}
}

@inproceedings{bender2021stochastic,
  title     = {On the dangers of stochastic parrots: Can language models be too big?},
  author    = {Bender, Emily M and Gebru, Timnit and McMillan-Major, Angelina and Shmitchell, Shmargaret},
  booktitle = {Proceedings of the 2021 ACM conference on fairness, accountability, and transparency},
  pages     = {610--623},
  year      = {2021}
}

@article{blodgett2020critical,
  title   = {Language (technology) is power: A critical survey of" bias" in nlp},
  author  = {Blodgett, Su Lin and Barocas, Solon and Daum{\'e} Iii, Hal and Wallach, Hanna},
  journal = {arXiv preprint arXiv:2005.14050},
  year    = {2020}
}

@article{bommasani2021foundation,
  title={On the opportunities and risks of foundation models},
  author={Bommasani, Rishi and Hudson, Drew A and Adeli, Ehsan and Altman, Russ and Arora, Simran and von Arx, Sydney and Bernstein, Michael S and Bohg, Jeannette and Bosselut, Antoine and Brunskill, Emma and others},
  journal={arXiv preprint arXiv:2108.07258},
  year={2021}
}

@article{bountos2023fomo,
  title   = {FoMo-Bench: a multi-modal, multi-scale and multi-task Forest Monitoring Benchmark for remote sensing foundation models},
  author  = {Bountos, Nikolaos Ioannis and Ouaknine, Arthur and Rolnick, David},
  journal = {arXiv preprint arXiv:2312.10114},
  year    = {2023}
}

@article{burke2021using,
  title     = {Using satellite imagery to understand and promote sustainable development},
  author    = {Burke, Marshall and Driscoll, Anne and Lobell, David B and Ermon, Stefano},
  journal   = {Science},
  volume    = {371},
  number    = {6535},
  pages     = {eabe8628},
  year      = {2021},
  publisher = {American Association for the Advancement of Science}
}

@article{chen2023palix,
  title   = {Pali-x: On scaling up a multilingual vision and language model},
  author  = {Chen, Xi and Djolonga, Josip and Padlewski, Piotr and Mustafa, Basil and Changpinyo, Soravit and Wu, Jialin and Ruiz, Carlos Riquelme and Goodman, Sebastian and Wang, Xiao and Tay, Yi and others},
  journal = {arXiv preprint arXiv:2305.18565},
  year    = {2023}
}

@article{chen2024internvl25,
  title   = {Expanding performance boundaries of open-source multimodal models with model, data, and test-time scaling},
  author  = {Chen, Zhe and Wang, Weiyun and Cao, Yue and Liu, Yangzhou and Gao, Zhangwei and Cui, Erfei and Zhu, Jinguo and Ye, Shenglong and Tian, Hao and Liu, Zhaoyang and others},
  journal = {arXiv preprint arXiv:2412.05271},
  year    = {2024}
}

@article{chen2025internvl3,
  title   = {Internvl3: Exploring advanced training and test-time recipes for open-source multimodal models},
  author  = {Zhu, Jinguo and Wang, Weiyun and Chen, Zhe and Liu, Zhaoyang and Ye, Shenglong and Gu, Lixin and Tian, Hao and Duan, Yuchen and Su, Weijie and Shao, Jie and others},
  journal = {arXiv preprint arXiv:2504.10479},
  year    = {2025}
}

@inproceedings{chuang2309decoding,
  title     = {DoLa: Decoding by Contrasting Layers Improves Factuality in Large Language Models},
  author    = {Chuang, Yung-Sung and Xie, Yujia and Luo, Hongyin and Kim, Yoon and Glass, James R and He, Pengcheng},
  booktitle = {The Twelfth International Conference on Learning Representations},
  year      = {2024}
}

@article{corsi2012demographic,
  title     = {Demographic and health surveys: a profile},
  author    = {Corsi, Daniel J and Neuman, Melissa and Finlay, Jocelyn E and Subramanian, SV},
  journal   = {International journal of epidemiology},
  volume    = {41},
  number    = {6},
  pages     = {1602--1613},
  year      = {2012},
  publisher = {Oxford University Press}
}

@article{dai2023instructblip,
  title   = {Instructblip: Towards general-purpose vision-language models with instruction tuning},
  author  = {Dai, Wenliang and Li, Junnan and Li, Dongxu and Tiong, Anthony and Zhao, Junqi and Wang, Weisheng and Li, Boyang and Fung, Pascale N and Hoi, Steven},
  journal = {Advances in neural information processing systems},
  volume  = {36},
  pages   = {49250--49267},
  year    = {2023}
}

@inproceedings{danish2025geobenchvlm,
  title     = {Geobench-vlm: Benchmarking vision-language models for geospatial tasks},
  author    = {Danish, Muhammad and Munir, Muhammad Akhtar and Shah, Syed Roshaan Ali and Kuckreja, Kartik and Khan, Fahad Shahbaz and Fraccaro, Paolo and Lacoste, Alexandre and Khan, Salman},
  booktitle = {Proceedings of the IEEE/CVF International Conference on Computer Vision},
  pages     = {7132--7142},
  year      = {2025}
}

@inproceedings{fu2023mme,
  title     = {Mme: A comprehensive evaluation benchmark for multimodal large language models},
  author    = {Fu, Chaoyou and Chen, Peixian and Shen, Yunhang and Qin, Yulei and Zhang, Mengdan and Lin, Xu and Yang, Jinrui and Zheng, Xiawu and Li, Ke and Sun, Xing and others},
  booktitle = {The Thirty-ninth Annual Conference on Neural Information Processing Systems Datasets and Benchmarks Track},
  year      = {2025}
}

@misc{google2024gemini2,
  title  = {Gemini 2.0},
  author = {{Google DeepMind}},
  year   = {2024},
  url    = {https://deepmind.google/technologies/gemini/}
}

@article{hardt2016equality,
  title={Equality of opportunity in supervised learning},
  author={Hardt, Moritz and Price, Eric and Srebro, Nati},
  journal={Advances in neural information processing systems},
  volume={29},
  year={2016}
}

@inproceedings{head2017human,
  title     = {Can human development be measured with satellite imagery?},
  author    = {Head, Andrew and Manguin, M{\'e}lanie and Tran, Nhat and Blumenstock, Joshua E},
  booktitle = {Proceedings of the Ninth International Conference on Information and Communication Technologies and Development},
  pages     = {1--11},
  year      = {2017}
}

@article{henderson2012measuring,
  title={Measuring economic growth from outer space},
  author={Henderson, J Vernon and Storeygard, Adam and Weil, David N},
  journal={American economic review},
  volume={102},
  number={2},
  pages={994--1028},
  year={2012},
  publisher={American Economic Association}
}

@inproceedings{huang-etal-2025-ai,
  title     = {Ai sees your location—but with a bias toward the wealthy world},
  author    = {Huang, Jingyuan and Huang, Jen-tse and Liu, Ziyi and Liu, Xiaoyuan and Wang, Wenxuan and Zhao, Jieyu},
  booktitle = {Proceedings of the 2025 Conference on Empirical Methods in Natural Language Processing},
  pages     = {18030--18050},
  year      = {2025}
}

@article{huang2025visbias,
  title   = {VisBias: Measuring Explicit and Implicit Social Biases in Vision Language Models},
  author  = {Huang, Jen-tse and Qin, Jiantong and Zhang, Jianping and Yuan, Youliang and Wang, Wenxuan and Zhao, Jieyu},
  journal = {arXiv preprint arXiv:2503.07575},
  year    = {2025}
}

@inproceedings{hudson2019gqa,
  title     = {Gqa: A new dataset for real-world visual reasoning and compositional question answering},
  author    = {Hudson, Drew A and Manning, Christopher D},
  booktitle = {Proceedings of the IEEE/CVF conference on computer vision and pattern recognition},
  pages     = {6700--6709},
  year      = {2019}
}

@inproceedings{janghorbani2023multi,
  title     = {Multi-modal bias: Introducing a framework for stereotypical bias assessment beyond gender and race in vision--language models},
  author    = {Janghorbani, Sepehr and De Melo, Gerard},
  booktitle = {Proceedings of the 17th Conference of the European Chapter of the Association for Computational Linguistics},
  pages     = {1725--1735},
  year      = {2023}
}

@article{jean2016combining,
  title   = {Combining satellite imagery and machine learning to predict poverty},
  author  = {Neal Jean and Marshall Burke and Michael Xie and W. Matthew Davis and David B. Lobell and Stefano Ermon},
  journal = {Science},
  volume  = {353},
  number  = {6301},
  pages   = {790--794},
  year    = {2016},
  doi     = {10.1126/science.aaf7894}
}

@inproceedings{kuckreja2024geochat,
  title     = {Geochat: Grounded large vision-language model for remote sensing},
  author    = {Kuckreja, Kartik and Danish, Muhammad Sohail and Naseer, Muzammal and Das, Abhijit and Khan, Salman and Khan, Fahad Shahbaz},
  booktitle = {Proceedings of the IEEE/CVF Conference on Computer Vision and Pattern Recognition},
  pages     = {27831--27840},
  year      = {2024}
}

@inproceedings{leng2024vcd,
  title     = {Mitigating object hallucinations in large vision-language models through visual contrastive decoding},
  author    = {Leng, Sicong and Zhang, Hang and Chen, Guanzheng and Li, Xin and Lu, Shijian and Miao, Chunyan and Bing, Lidong},
  booktitle = {Proceedings of the IEEE/CVF Conference on Computer Vision and Pattern Recognition},
  pages     = {13872--13882},
  year      = {2024}
}

@inproceedings{li2023blip,
  title        = {Blip-2: Bootstrapping language-image pre-training with frozen image encoders and large language models},
  author       = {Li, Junnan and Li, Dongxu and Savarese, Silvio and Hoi, Steven},
  booktitle    = {International conference on machine learning},
  pages        = {19730--19742},
  year         = {2023},
  organization = {PMLR}
}

@inproceedings{li2023contrastive,
  title     = {Contrastive decoding: Open-ended text generation as optimization},
  author    = {Li, Xiang Lisa and Holtzman, Ari and Fried, Daniel and Liang, Percy and Eisner, Jason and Hashimoto, Tatsunori B and Zettlemoyer, Luke and Lewis, Mike},
  booktitle = {Proceedings of the 61st annual meeting of the association for computational linguistics (volume 1: Long papers)},
  pages     = {12286--12312},
  year      = {2023}
}

@article{li2023pope,
  title   = {Evaluating object hallucination in large vision-language models},
  author  = {Li, Yifan and Du, Yifan and Zhou, Kun and Wang, Jinpeng and Zhao, Wayne Xin and Wen, Ji-Rong},
  journal = {arXiv preprint arXiv:2305.10355},
  year    = {2023}
}

@article{li2024llavaonevision,
  title   = {Llava-onevision: Easy visual task transfer},
  author  = {Li, Bo and Zhang, Yuanhan and Guo, Dong and Zhang, Renrui and Li, Feng and Zhang, Hao and Zhang, Kaichen and Zhang, Peiyuan and Li, Yanwei and Liu, Ziwei and others},
  journal = {arXiv preprint arXiv:2408.03326},
  year    = {2024}
}

@article{lin2024languagepriors,
  title   = {Revisiting the role of language priors in vision-language models},
  author  = {Lin, Zhiqiu and Chen, Xinyue and Pathak, Deepak and Zhang, Pengchuan and Ramanan, Deva},
  journal = {arXiv preprint arXiv:2306.01879},
  year    = {2023}
}

@article{liu2023llava,
  title   = {Visual instruction tuning},
  author  = {Liu, Haotian and Li, Chunyuan and Wu, Qingyang and Lee, Yong Jae},
  journal = {Advances in neural information processing systems},
  volume  = {36},
  pages   = {34892--34916},
  year    = {2023}
}

@inproceedings{liu2024llava15,
  title     = {Improved baselines with visual instruction tuning},
  author    = {Liu, Haotian and Li, Chunyuan and Li, Yuheng and Lee, Yong Jae},
  booktitle = {Proceedings of the IEEE/CVF conference on computer vision and pattern recognition},
  pages     = {26296--26306},
  year      = {2024}
}

@article{lu2022scienceqa,
  title   = {Learn to explain: Multimodal reasoning via thought chains for science question answering},
  author  = {Lu, Pan and Mishra, Swaroop and Xia, Tanglin and Qiu, Liang and Chang, Kai-Wei and Zhu, Song-Chun and Tafjord, Oyvind and Clark, Peter and Kalyan, Ashwin},
  journal = {Advances in Neural Information Processing Systems},
  volume  = {35},
  pages   = {2507--2521},
  year    = {2022}
}

@article{manvi2024geobias,
  title   = {Large language models are geographically biased},
  author  = {Manvi, Rohin and Khanna, Samar and Burke, Marshall and Lobell, David and Ermon, Stefano},
  journal = {arXiv preprint arXiv:2402.02680},
  year    = {2024}
}

@article{marsocci2024pangaea,
  title   = {Pangaea: A global and inclusive benchmark for geospatial foundation models},
  author  = {Marsocci, Valerio and Jia, Yuru and Bellier, Georges Le and Kerekes, David and Zeng, Liang and Hafner, Sebastian and Gerard, Sebastian and Brune, Eric and Yadav, Ritu and Shibli, Ali and others},
  journal = {arXiv preprint arXiv:2412.04204},
  year    = {2024}
}

@inproceedings{mitchell2019model,
  title     = {Model cards for model reporting},
  author    = {Mitchell, Margaret and Wu, Simone and Zaldivar, Andrew and Barnes, Parker and Vasserman, Lucy and Hutchinson, Ben and Spitzer, Elena and Raji, Inioluwa Deborah and Gebru, Timnit},
  booktitle = {Proceedings of the conference on fairness, accountability, and transparency},
  pages     = {220--229},
  year      = {2019}
}

@inproceedings{nadeem2021stereoset,
  title     = {StereoSet: Measuring stereotypical bias in pretrained language models},
  author    = {Nadeem, Moin and Bethke, Anna and Reddy, Siva},
  booktitle = {Proceedings of the 59th annual meeting of the association for computational linguistics and the 11th international joint conference on natural language processing (volume 1: long papers)},
  pages     = {5356--5371},
  year      = {2021}
}

@inproceedings{nangia2020crows,
  title     = {CrowS-pairs: A challenge dataset for measuring social biases in masked language models},
  author    = {Nangia, Nikita and Vania, Clara and Bhalerao, Rasika and Bowman, Samuel},
  booktitle = {Proceedings of the 2020 conference on empirical methods in natural language processing (EMNLP)},
  pages     = {1953--1967},
  year      = {2020}
}

@inproceedings{niu2021counterfactual,
  title     = {Counterfactual vqa: A cause-effect look at language bias},
  author    = {Niu, Yulei and Tang, Kaihua and Zhang, Hanwang and Lu, Zhiwu and Hua, Xian-Sheng and Wen, Ji-Rong},
  booktitle = {Proceedings of the IEEE/CVF conference on computer vision and pattern recognition},
  pages     = {12700--12710},
  year      = {2021}
}

@misc{openai2024gpt4o,
  title  = {Hello GPT-4o},
  author = {{OpenAI}},
  year   = {2024},
  url    = {https://openai.com/index/hello-gpt-4o/},
  note   = {Accessed: 2024-05-13}
}

@article{pouget2024nofilter,
  title   = {No filter: Cultural and socioeconomic diversity in contrastive vision-language models},
  author  = {Pouget, Ang{\'e}line and Beyer, Lucas and Bugliarello, Emanuele and Wang, Xiao and Steiner, Andreas and Zhai, Xiaohua and Alabdulmohsin, Ibrahim M},
  journal = {Advances in Neural Information Processing Systems},
  volume  = {37},
  pages   = {106474--106496},
  year    = {2024}
}

@inproceedings{radford2021learning,
  title        = {Learning transferable visual models from natural language supervision},
  author       = {Radford, Alec and Kim, Jong Wook and Hallacy, Chris and Ramesh, Aditya and Goh, Gabriel and Agarwal, Sandhini and Sastry, Girish and Askell, Amanda and Mishkin, Pamela and Clark, Jack and others},
  booktitle    = {International conference on machine learning},
  pages        = {8748--8763},
  year         = {2021},
  organization = {PmLR}
}

@inproceedings{ross2021groundedbias,
  title     = {Measuring social biases in grounded vision and language embeddings},
  author    = {Ross, Candace and Katz, Boris and Barbu, Andrei},
  booktitle = {Proceedings of the 2021 Conference of the North American Chapter of the Association for Computational Linguistics: Human Language Technologies},
  pages     = {998--1008},
  year      = {2021}
}

@incollection{ruggeri2023multidimensional,
  title     = {A multi-dimensional study on bias in vision-language models},
  author    = {Ruggeri, Gabriele and Nozza, Debora and others},
  booktitle = {Findings of the Association for Computational Linguistics: ACL 2023},
  year      = {2023},
  publisher = {Association for Computational Linguistics}
}

@inproceedings{soni2025earthdial,
  title     = {Earthdial: Turning multi-sensory earth observations to interactive dialogues},
  author    = {Soni, Sagar and Dudhane, Akshay and Debary, Hiyam and Fiaz, Mustansar and Munir, Muhammad Akhtar and Danish, Muhammad Sohail and Fraccaro, Paolo and Watson, Campbell D and Klein, Levente J and Khan, Fahad Shahbaz and others},
  booktitle = {Proceedings of the Computer Vision and Pattern Recognition Conference},
  pages     = {14303--14313},
  year      = {2025}
}

@misc{un2030agenda2015,
  author       = {{United Nations General Assembly}},
  title        = {Transforming our world: the 2030 Agenda for Sustainable Development},
  year         = {2015},
  howpublished = {Resolution A/RES/70/1},
  url          = {https://undocs.org/A/RES/70/1},
  note         = {Adopted on 25 September 2015}
}

@article{wang2024icd,
  title   = {Mitigating hallucinations in large vision-language models with instruction contrastive decoding},
  author  = {Wang, Xintong and Pan, Jingheng and Ding, Liang and Biemann, Chris},
  journal = {arXiv preprint arXiv:2403.18715},
  year    = {2024}
}

@inproceedings{weng2024images,
  title     = {Images speak louder than words: Understanding and mitigating bias in vision-language model from a causal mediation perspective},
  author    = {Weng, Zhaotian and Gao, Zijun and Andrews, Jerone and Zhao, Jieyu},
  booktitle = {Proceedings of the 2024 Conference on Empirical Methods in Natural Language Processing},
  pages     = {15669--15680},
  year      = {2024}
}

@article{weng2024imagesspeak,
  title   = {Images Speak Louder than Words: Understanding and Mitigating Bias in Vision-Language Model from a Causal Mediation Perspective},
  author  = {Weng, Zhaotian and Gao, Zijun and Andrews, Jerone and Zhao, Jieyu},
  journal = {arXiv preprint arXiv:2407.02814},
  year    = {2024}
}

@article{yeh2020publicly,
  title     = {Using publicly available satellite imagery and deep learning to understand economic well-being in Africa},
  author    = {Yeh, Christopher and Perez, Anthony and Driscoll, Anne and Azzari, George and Tang, Zhongyi and Lobell, David and Ermon, Stefano and Burke, Marshall},
  journal   = {Nature communications},
  volume    = {11},
  number    = {1},
  pages     = {2583},
  year      = {2020},
  publisher = {Nature Publishing Group UK London}
}

@article{yeh2021sustainbench,
  title   = {Sustainbench: Benchmarks for monitoring the sustainable development goals with machine learning},
  author  = {Yeh, Christopher and Meng, Chenlin and Wang, Sherrie and Driscoll, Anne and Rozi, Erik and Liu, Patrick and Lee, Jihyeon and Burke, Marshall and Lobell, David B and Ermon, Stefano},
  journal = {arXiv preprint arXiv:2111.04724},
  year    = {2021}
}

@inproceedings{zhang2022counterbias,
  title     = {Counterfactually measuring and eliminating social bias in vision-language pre-training models},
  author    = {Zhang, Yi and Wang, Junyang and Sang, Jitao},
  booktitle = {Proceedings of the 30th ACM International Conference on Multimedia},
  pages     = {4996--5004},
  year      = {2022}
}

@inproceedings{zhang2025debiasing,
  title     = {Debiasing multimodal large language models via penalization of language priors},
  author    = {Zhang, YiFan and Shi, Yang and Yu, Weichen and Wen, Qingsong and Wang, Xue and Yang, Wenjing and Zhang, Zhang and Wang, Liang and Jin, Rong},
  booktitle = {Proceedings of the 33rd ACM International Conference on Multimedia},
  pages     = {4232--4241},
  year      = {2025}
}

@inproceedings{zhou2022vlstereoset,
  title     = {Vlstereoset: A study of stereotypical bias in pre-trained vision-language models},
  author    = {Zhou, Kankan and Lai, Eason and Jiang, Jing},
  booktitle = {Proceedings of the 2nd Conference of the Asia-Pacific Chapter of the Association for Computational Linguistics and the 12th International Joint Conference on Natural Language Processing (Volume 1: Long Papers)},
  pages     = {527--538},
  year      = {2022}
}

@inproceedings{zhu2025ibd,
  title     = {Ibd: Alleviating hallucinations in large vision-language models via image-biased decoding},
  author    = {Zhu, Lanyun and Ji, Deyi and Chen, Tianrun and Xu, Peng and Ye, Jieping and Liu, Jun},
  booktitle = {Proceedings of the Computer Vision and Pattern Recognition Conference},
  pages     = {1624--1633},
  year      = {2025}
}
}

\clearpage \appendix

\setcounter{section}{0}
\setcounter{table}{0}
\setcounter{figure}{0}
\setcounter{equation}{0}

\twocolumn[{
\renewcommand\twocolumn[1][]{#1}
\nolinenumbers
\begin{center}
\vspace{-3pt}
\Large \textbf{SDGBiasBench: Benchmarking and Mitigating Vision–Language Models' Biases in Sustainable Development Goals} \\
\Large \vspace{0.3em} Supplementary Material \\
\end{center}
\ifarxiv \linenumbers \fi
\vspace{1.0em}
}]

\pagenumbering{gobble}
\begin{strip}
    \vspace{-2cm}
    \begin{spacing}{0.94}
        \normalsize \tableofcontents
    \end{spacing}
\end{strip}
\newpage
\ifarxiv
\pagenumbering{arabic}
\setcounter{page}{13}
\fi

\renewcommand{\thesection}{\Alph{section}}
\renewcommand{\thetable}{\Alph{table}}
\renewcommand{\thefigure}{\Alph{figure}}
\renewcommand{\theequation}{\Alph{equation}}

\section{SDGBiasBench: A Comprehensive Description}
\label{sec:appendix_sdgbiasbench}

\subsection{Overview}
SDGBiasBench evaluates vision--language reasoning on real-world SDG-related indicators while revealing SDG biases. It comprises two task types: multiple-choice questions (MCQs) and regression tasks to mirror practical use cases. Each datapoint is paired with satellite imagery and often augmented with street-level photos; each sample also includes structured context variables (\eg sanitation index, women’s education level, asset index), encouraging multi-step reasoning across modalities. The benchmark draws from raw data in \textsc{SustainBench}~\citep{yeh2021sustainbench} and the Demographic and Health Surveys (DHS)~\citep{corsi2012demographic}. We extend them through human-authored high-level reasoning questions and expert-labeled ground truth answers. Below we summarize the three pillars in detail.

\paragraph{Key features.}
\begin{itemize}
  \item \textbf{Diverse questions.} SDGBiasBench contains 500,000 multiple-choice questions and 50,000 regression queries targeting SDG-related indicators (\eg poverty, infrastructure access, educational attainment). Together, MCQ and regression tasks reflect real SDG practice, requiring both discrete judgments and accurate prediction of continuous development indicators.
  \item \textbf{Multi-modal context.} Each task is paired with satellite imagery, often complemented by street-level photos, and structured socio-economic context variables (\eg sanitation index, women’s education, asset index), requiring joint reasoning over visual and textual modalities.
  \item \textbf{Sustainability pillars.} The benchmark covers three sustainability pillars---Health \& Nutrition, Basic Services \& Infrastructure, and Human Capital \& Development---to support broad assessment across SDG themes.
\end{itemize}

\paragraph{Design goal.}
Overall, SDGBiasBench is structured to move beyond surface perception: each task encourages models to synthesize signals across modalities and to revise judgments under changing evidence, making it well-suited for diagnosing SDG-specific prior reliance and directional errors.

\subsection{Dataset Statistics and Annotation Protocol}
\label{sec:dataset_stat_and_annotation}
We summarize the benchmark composition and per-pillar MCQ label distributions in~\Cref{tab:appendix_pillar_counts,tab:appendix_pillar_label_dist}, respectively. SDGBiasBench contains both multiple-choice questions (MCQs) and regression tasks across three sustainability pillars. 

\begin{table}[t]
\centering
\caption{\textbf{Number of questions in SDGBiasBench by pillar and task type.}}
\label{tab:appendix_pillar_counts}
\small
\begin{tabular}{p{0.52\columnwidth}cc}
\toprule
\textbf{Dataset Pillar} & \textbf{MCQ} & \textbf{Regression} \\
\midrule
Health \& Nutrition & 222,900 & 19,427 \\
Basic Services \& Infrastructure & 162,600 & 16,761 \\
Human Capital \& Development & 114,500 & 13,812 \\
\midrule
\textbf{Total} & \textbf{500,000} & \textbf{50,000} \\
\bottomrule
\end{tabular}
\end{table}

\begin{table}[!t]
\centering
\caption{\textbf{MCQ label distributions (\%) by pillar.}}
\label{tab:appendix_pillar_label_dist}
\small
\begin{tabular}{p{0.52\columnwidth}ccc}
\toprule
\textbf{Dataset Pillar} & \textbf{A} & \textbf{B} & \textbf{C} \\
\midrule
Health \& Nutrition & 27.75 & 42.80 & 29.45 \\
Basic Services \& Infrastructure & 40.11 & 39.11 & 20.78 \\
Human Capital \& Development & 32.03 & 33.87 & 34.11 \\
\bottomrule
\end{tabular}
\end{table}

\paragraph{Construction and annotation pipeline.}
SDGBiasBench is constructed through a structured expert-involved pipeline to ensure that each example is geographically grounded, semantically meaningful, and quality-controlled. 
We provide more details on the pipeline below:

\begin{itemize}
    \item \textbf{Source grounding and alignment.} Each example is anchored to a geo-entity. We align satellite imagery, street-level imagery, and structured context variables from \textsc{SustainBench} and Demographic and Health Surveys (DHS) using location identifiers, so that all available modalities correspond to the same region.
    \item \textbf{Expert-defined indicator templates.} We invite 9 sustainability-domain experts to define the question templates for each indicator, together with the required evidence sources, option-bin rules, and intended reasoning targets. For each task, the annotation protocol specifies which modalities are needed, what latent construct should be inferred, and how labels or regression targets should be interpreted.
    \item \textbf{Option and bin construction.} For MCQs, we discretize each indicator into ordinal answer bins using either established SDG-related thresholds or indicator-specific distributional cut points. The resulting options are mutually exclusive and semantically ordered sub-ranges of the underlying indicator, enabling qualitative judgment over value ranges rather than exact numerical prediction. For regression tasks, the targets are continuous ground-truth values obtained directly from \textsc{SustainBench} and Demographic and Health Surveys (DHS).
    \item \textbf{Quality control.} We apply both automatic and human quality-control procedures. Automatic checks remove examples with missing modalities, inconsistent metadata, duplicated or near-duplicate items, and invalid option sets. Human quality assurance includes template review and stratified spot-checking across pillars and task types to verify semantic correctness, answer validity, and consistency of evidence alignment. In particular, 3 experts are responsible for this quality-control stage.
    \item \textbf{Human review protocol.} A stratified subset of approximately 5\% of the benchmark is human-reviewed across pillars and task types. Each reviewed item is independently checked by two experts, and disagreements are resolved by the third expert. This process validates the expert-defined templates, option construction, answer validity, and modality alignment, while the full benchmark is generated from the validated rules and aligned ground-truth indicators.
\end{itemize}

\subsection{Pillar 1: Health \& Nutrition}
This pillar evaluates whether VLMs can reason about population health, nutritional status, and healthcare access from indirect, multi-modal evidence. Tasks span child malnutrition, disease burden, physical activity, and health system quality, requiring models to integrate visual cues (\eg settlement density, environmental conditions, infrastructure) with structured socio-economic context. Crucially, both qualitative judgments and quantitative estimation settings emphasize \emph{latent health inference}: target indicators cannot be directly observed from a single modality, but must be inferred through correlated proxies. This pillar therefore probes a model’s ability to move beyond surface correlations and performs multi-step reasoning about human well-beings.

\paragraph{MCQ tasks.} We define five MCQs:
\begin{itemize}
  \item Q1: Stunting in children under 5 (SDG 2.2.1)
  \item Q2: Universal Health Coverage index (SDG 3.8.1)
  \item Q3: Insufficient physical activity prevalence (WHO)
  \item Q4: Wasting in children under 5 (SDG 2.2.2)
  \item Q5: Healthcare Access and Quality (HAQ) Index (IHME)
\end{itemize}
\emph{All MCQs ask models to infer the regional indicator by jointly using imagery and structured context.}

\paragraph{Regression tasks.} We include four regression questions:
\begin{itemize}
  \item Q1: Survival Infrastructure (\emph{infer} Under-5 mortality rate)
  \item Q2: Maternal Capital (\emph{infer} Under-5 mortality rate)
  \item Q3: Nutritional Wealth (\emph{infer} Women’s BMI)
  \item Q4: Total Development (\emph{infer} Women’s BMI)
\end{itemize}
\emph{These regressions require continuous inference from indirect evidence rather than reading the target directly from any single modality.}

\subsection{Pillar 2: Basic Services \& Infrastructure}
This pillar focuses on access to essential services such as electricity, water, and sanitation, as well as their downstream health implications. Tasks assess whether models can connect visible infrastructure patterns such as roads, buildings, utilities, and environmental features with structured indices which capture service coverage and deprivation. Beyond direct access estimation, several tasks require \emph{inverse reconstruction}: inferring latent socio-economic status or infrastructure quality from human activity patterns and environmental signals. As such, this pillar stresses spatial reasoning, cross-modal alignment, and the ability to recover unobserved service conditions from partial evidence.

\paragraph{MCQ tasks.} Four MCQs cover:
\begin{itemize}
  \item Q1: Electricity access (SDG 7.1.1)
  \item Q2: Access to safe drinking water (SDG 6.1.1)
  \item Q3: Access to sanitation services (SDG 6.2.1)
  \item Q4: WASH-attributable mortality rate (SDG 3.9.2)
\end{itemize}
\emph{Questions test whether models can connect visible infrastructure cues with structured indices to estimate access or burden.}

\paragraph{Regression tasks.} Four regression questions target service/wealth reconstruction:
\begin{itemize}
  \item Q1: Hidden Poverty (\emph{infer} Asset index)
  \item Q2: Plumbing from People (\emph{infer} Sanitation index)
  \item Q3: Water Security (\emph{infer} Water index)
  \item Q4: Sanitation Reconstruction (\emph{infer} Sanitation index)
\end{itemize}
\emph{These tasks emphasize inverse reasoning---recovering infrastructure or SES from human and environmental signals.}

\begin{table}[t]
    \centering
    \caption{\textbf{MCQ prompt template.} The \textbf{Context} line is included only when structured SDG variables are provided.}
    \small
    \setlength{\tabcolsep}{6pt}
    \renewcommand{\arraystretch}{1.12}
    \begin{tabular}{p{0.97\columnwidth}}
        \toprule
        \textbf{Prompt (MCQ)} \\
        \midrule
        \textbf{Question:} \textcolor{magenta}{\{QUESTION\}} \\
        \textbf{Context:} \textcolor{magenta}{\{CONTEXT\}} \ \emph{(optional)} \\
        \textbf{Options:} \textcolor{magenta}{\{OPTIONS\}} \\
        \texttt{VALID OPTIONS: \{VALID\_LETTERS\}} \\
        \texttt{STRICT OUTPUT: Return EXACTLY ONE uppercase letter from VALID OPTIONS.} \\
        \texttt{Output nothing else (no words, numbers, symbols, spaces, or newlines).} \\
        \bottomrule
    \end{tabular}
    \label{tab:prompt-mcq}
\end{table}

\begin{table}[t]
    \centering
    \caption{\textbf{Regression prompt template.} The \textbf{Context} line is included only when structured SDG variables are provided.}
    \small
    \setlength{\tabcolsep}{6pt}
    \renewcommand{\arraystretch}{1.12}
    \begin{tabular}{p{0.97\columnwidth}}
        \toprule
        \textbf{Prompt (Regression)} \\
        \midrule
        \textbf{Question:} \textcolor{magenta}{\{QUESTION\}} \\
        \textbf{Context:} \textcolor{magenta}{\{CONTEXT\}} \ \emph{(optional)} \\
        \texttt{Answer strictly with a single numerical value. Round to 1 decimal place.} \\
        \bottomrule
    \end{tabular}
    \label{tab:prompt-reg}
\end{table}

\begin{table}[t]
\centering
\caption{\textbf{Operational definition of \textbf{T1--T5} context conditions} used in Table~\ref{tab:mcq-results}.}
\small
\setlength{\tabcolsep}{4pt}
\renewcommand{\arraystretch}{1.1}
\begin{tabularx}{\linewidth}{
>{\bfseries}p{0.10\linewidth}
>{\raggedright\arraybackslash}p{0.3\linewidth}
>{\raggedright\arraybackslash}X}
\toprule
\textbf{Test} & \textbf{Context form} & \textbf{Example} \\
\midrule
T1 & Value-given (full) &
\textit{Given:} asset\_index=1.0, water\_index=3.4, sanitation\_index=2.6, under\_5\_mortality\_rate=0, woman\_education\_years=9.5, woman\_bmi=24.4, urban=Urban \\
T2 & Value-given (reduced) &
\textit{Given:} asset\_index=1.0, sanitation\_index=2.6, urban=Urban \\
T3 & Name-only hint (full) &
\textit{You may consider:} asset\_index, water\_index, sanitation\_index, under\_5\_mortality\_rate, woman\_education\_years, woman\_bmi, urban \\
T4 & Name-only hint (reduced) &
\textit{You may consider:} asset\_index, sanitation\_index, urban \\
T5 & No context &
Question only (no structured variables provided or hinted). \\
\bottomrule
\end{tabularx}
\label{tab:t1t5_definition}
\end{table}

\subsection{Pillar 3: Human Capital \& Development}
This pillar examines reasoning about education and broader human development, where signals are diffuse and highly correlated with other aspects of regional context. Models must infer educational attainment and composite development indices by synthesizing multi-modal proxies such as settlement structure, asset distribution, and service availability. The emphasis is on \emph{abstract human-capital inference} such as human development index and education–health alignment. Compared to the other pillars, this setting places greater weight on long-range correlations and holistic interpretation of socio-economic environments.

\paragraph{MCQ tasks.} Three MCQs include:
\begin{itemize}
  \item Q1: Female lower-secondary school completion (SDG 4.1.2)
  \item Q2: Subnational Human Development Index (SHDI) (UNDP)
  \item Q3: Education--Health alignment score (custom indicator)
\end{itemize}
\emph{Models must map multi-modal proxies (settlement structure, assets, services) to human-capital indicators.}

\paragraph{Regression task.} One regression question:
\begin{itemize}
  \item Q1: Education Access (\emph{infer} Women’s education level)
\end{itemize}
\emph{This task reconstructs schooling from correlated wealth/health/service cues, making it a non-trivial, indirect inference.}

\subsection{Prompt Templates}
For all models and evidence views, we construct the textual prompt by concatenating the \emph{question} with (optionally) the structured SDG \emph{context}, followed by a task-specific output constraint.
For multiple-choice questions (MCQ), we additionally provide the option set and enforce a \emph{single-letter} response (Table~\ref{tab:prompt-mcq}).
For regression queries, we enforce a \emph{single numeric} response rounded to one decimal place (Table~\ref{tab:prompt-reg}).
When structured context is unavailable (\eg question-only or image-only views), the \textbf{Context} line is omitted.

\begin{table}[!t]
\centering
\caption{\textbf{Tolerances used to compute regression interval accuracy.}}
\small
\setlength{\tabcolsep}{7pt}
\renewcommand{\arraystretch}{1.15}
\begin{tabular}{lc}
\toprule
\textbf{Indicator} & \textbf{Tolerance $\delta_k$} \\
\midrule
under 5 mortality rate & $\pm~5.0$ \\
women bmi & $\pm~1.0$ \\
asset index & $\pm~0.75$ \\
sanitation index & $\pm~0.75$ \\
water index & $\pm~0.75$ \\
women edu & $\pm~1.5$ \\
\bottomrule
\end{tabular}
\label{tab:reg_interval_tolerance}
\end{table}
\begin{table}[t]
\centering
\caption{\textbf{Hyperparameter-search ranges for CADE.}}
\small
\setlength{\tabcolsep}{7pt}
\renewcommand{\arraystretch}{1.15}
\begin{tabular}{lcc}
\toprule
\textbf{Hyperparameter} & \textbf{MCQ range} & \textbf{Regression range} \\
\midrule
$\alpha$ & $[0.5,\,2.0]$ & $[0.0,\,7.0]$ \\
$\lambda_{\mathrm{KL}}$ & $[0.0,\,5.0]$ & $[0.0,\,7.0]$ \\
$\beta$ & $[0.0,\,1.5]$ & $[0.0,\,7.0]$ \\
$\tau$ & $[0.65,\,0.99]$ & $[0.25,\,0.99]$ \\
\bottomrule
\end{tabular}
\label{tab:cade_hparam_ranges}
\end{table}

\section{Additional Implementation Details}
\label{sec:appendix_cade_details}

We provide further implementation details to support the reproducibility of our proposed benchmark with experimental setups below.

\subsection{Details of Five Context Tests (T1--T5) for MCQ}
\label{sec:t1_t5_explain}
We evaluate each multiple-choice question by transforming a single data instance into five context-test variants (T1--T5), which evaluate model's behavior under progressively weaker structured-context conditions. They are designed along two complementary axes: \textbf{context explicitness} and \textbf{context quantity}. In terms of explicitness, the tests span three levels: \textbf{value-given}, \textbf{name-only hint}, and \textbf{no context}. In terms of quantity, whenever structured context is present, we further distinguish between a \textbf{full} set of variables and a \textbf{reduced} subset. Accordingly, \textbf{T1} and \textbf{T2} provide exact structured variable values, with T1 using the full set and T2 using a reduced subset. \textbf{T3} and \textbf{T4} provide only variable names as soft guidance, again with T3 using the full set and T4 using a reduced subset. Finally, \textbf{T5} removes structured context entirely, leaving the model with the question alone. This design enables us to disentangle the effect of \textbf{how much} structured context is available (full \vs reduced) from \textbf{how explicitly} it is presented (value-given \vs name-only \vs none). ~\cref{tab:t1t5_definition} summarizes the operational definitions and representative examples of all five tests.

\paragraph{Remark.}
All reported performance in empirical analysis in~\cref{sec:sdg_bias_analysis} and ablation studies in~\cref{subsec:exp_ablation_sensitivity} are obtained only under context test T1 unless otherwise stated. 
Instead, average accuracy of the five tests for MCQ tasks are reported in~\cref{fig:mcq_mae_results_bar}.

\subsection{Details of Regression Variant}
For regression queries, we apply CADE to the \emph{first generated token} only. This design matches the next-token decoding mechanism of current VLMs: once the initial digit is determined, debiasing later tokens is not meaningful because subsequent generation is already conditioned on the first-token choice. Accordingly, we unify the candidate definition across MCQ and regression by specifying $\mathcal{O}_i$ as the option set: for MCQ, $\mathcal{O}_i$ is the set of answer choices (\eg $\{A,B,C\}$); for regression, we discretize the first-token space as $\mathcal{O}_i=\{0,1,\dots,9\}$, where $z^{(T)}_{i,c}$ denotes the logit that the first generated token is digit $o$ under view $T$. All subsequent steps (view-wise probabilities calculation, confidence-gated threshold, three streams engagement, contrastive scoring with adaptive disagreement) follow the same formulation as in the MCQ case.

\subsection{Details of Interval Accuracy Tolerances}
\label{sec:appendix_interval_accuracy}
We compute interval accuracy using pre-defined indicator-specific tolerances $\delta_k$,and a prediction $\hat{y}$ is counted correct if $|\hat{y}-y|\le \delta_k$.
The used tolerances for each indicator are given in~\Cref{tab:reg_interval_tolerance}.

\subsection{Details of Hyperparameter Search}
\paragraph{Hyperparameter ranges.}
CADE introduces a confidence gate $\tau$, an adaptive exponent base $\alpha$, a disagreement multiplier $\lambda_{\mathrm{KL}}$, and a prior-penalty coefficient $\beta$ (\cref{eq:cade-alpha}, \cref{eq:cade-score}). The random-search ranges are listed in~\cref{tab:cade_hparam_ranges}.

\vspace{-4pt}
\paragraph{Implementation details.}
We tune CADE via random search on a held-out validation split: we sample $10{,}000$ hyperparameter configurations (uniformly within preset ranges) and evaluate each configuration once on $20\%$ of the data.  We then re-evaluate the top 100 performing configuration on the full validation set to finalize the hyperparameters used in all reported results. Since all hyperparameters are sampled jointly in each trial, any given pair of hyperparameters is effectively explored through the same $10{,}000$ joint samples (\ie there is no separate pairwise grid sweep).

\section{Additional Experimental Results}
\label{sec:appendix_regression_results}

We provide comprehensive numerical results as the extension of~\Cref{fig:mcq_mae_results_bar}.
Specifically, we report Accuracy across all five context tests (T1--T5) for MCQ in~\Cref{tab:mcq-results}, and report MAE with Interval Accuracy for Regression in~\Cref{tab:regression-results}.

\setlength{\dbltextfloatsep}{6pt}
\begin{table*}[t]
\centering
\caption{\textbf{Performance of vision--language models on multiple-choice questions.} "Ours" refers to applying CADE to the same base model, where per-test gains are highlighted with green.}
\vspace{-3pt}
\label{tab:mcq-results}
\small
\setlength{\tabcolsep}{4pt}
\renewcommand{\arraystretch}{1.15}

\begin{tabularx}{\textwidth}{@{} l c c *{7}{>{\centering\arraybackslash}X} @{}}
\toprule
\textbf{Model} & \textbf{Params} & \textbf{Setting} &
\textbf{T1} & \textbf{T2} & \textbf{T3} & \textbf{T4} & \textbf{T5} & 
\textbf{Avg. Gain} & \textbf{Rel. Gain} \\
\midrule

\multicolumn{10}{@{}c@{}}{\textbf{Open-source VLMs}} \\
\midrule

\multirow{4}{*}{\textsc{LLaVA-v1.5}}
& \multirow{2}{*}{7B}
& Original & 0.30 & 0.32 & 0.22 & 0.26 & 0.29 & -- & -- \\
&  & Ours
& 0.53\rlap{\textsuperscript{\cgaphl{+}{0.23}}} & 0.54\rlap{\textsuperscript{\cgaphl{+}{0.22}}} & 0.55\rlap{\textsuperscript{\cgaphl{+}{0.33}}} & 0.54\rlap{\textsuperscript{\cgaphl{+}{0.28}}} & 0.49\rlap{\textsuperscript{\cgaphl{+}{0.20}}} & \textbf{\textcolor{nicergreen}{+0.25}} & \textbf{\textcolor{nicergreen}{+90.6\%}} \\
\cmidrule(lr){2-10}
& \multirow{2}{*}{13B}
& Original & 0.42 & 0.40 & 0.42 & 0.44 & 0.43 & -- & -- \\
&  & Ours 
& 0.57\rlap{\textsuperscript{\cgaphl{+}{0.15}}} & 0.48\rlap{\textsuperscript{\cgaphl{+}{0.08}}} & 0.52\rlap{\textsuperscript{\cgaphl{+}{0.10}}} & 0.51\rlap{\textsuperscript{\cgaphl{+}{0.07}}} & 0.50\rlap{\textsuperscript{\cgaphl{+}{0.07}}} & \textbf{\textcolor{nicergreen}{+0.09}} & \textbf{\textcolor{nicergreen}{+22.3\%}} \\
\midrule

\multirow{4}{*}{\textsc{InstructBLIP}}
& \multirow{2}{*}{7B}
& Original & 0.31 & 0.42 & 0.32 & 0.39 & 0.32 & -- & -- \\
&  & Ours
& 0.47\rlap{\textsuperscript{\cgaphl{+}{0.16}}} & 0.63\rlap{\textsuperscript{\cgaphl{+}{0.21}}} & 0.57\rlap{\textsuperscript{\cgaphl{+}{0.25}}} & 0.45\rlap{\textsuperscript{\cgaphl{+}{0.06}}} & 0.46\rlap{\textsuperscript{\cgaphl{+}{0.14}}} & \textbf{\textcolor{nicergreen}{+0.16}} & \textbf{\textcolor{nicergreen}{+46.6\%}} \\
\cmidrule(lr){2-10}
& \multirow{2}{*}{13B}
& Original & 0.34 & 0.48 & 0.33 & 0.34 & 0.34 & -- & -- \\
&  & Ours 
& 0.39\rlap{\textsuperscript{\cgaphl{+}{0.05}}} & 0.49\rlap{\textsuperscript{\cgaphl{+}{0.01}}} & 0.38\rlap{\textsuperscript{\cgaphl{+}{0.05}}} & 0.45\rlap{\textsuperscript{\cgaphl{+}{0.11}}} & 0.38\rlap{\textsuperscript{\cgaphl{+}{0.04}}} & \textbf{\textcolor{nicergreen}{+0.05}} & \textbf{\textcolor{nicergreen}{+14.2\%}} \\
\midrule

\multirow{2}{*}{\textsc{LLaVA-Onevision}}
& \multirow{2}{*}{7B}
& Original & 0.49 & 0.49 & 0.21 & 0.43 & 0.34 & -- & -- \\
&  & Ours
& 0.53\rlap{\textsuperscript{\cgaphl{+}{0.04}}} & 0.52\rlap{\textsuperscript{\cgaphl{+}{0.03}}} & 0.42\rlap{\textsuperscript{\cgaphl{+}{0.21}}} & 0.48\rlap{\textsuperscript{\cgaphl{+}{0.05}}} & 0.40\rlap{\textsuperscript{\cgaphl{+}{0.06}}} & \textbf{\textcolor{nicergreen}{+0.08}} & \textbf{\textcolor{nicergreen}{+19.9\%}} \\
\midrule

\multirow{4}{*}{\textsc{Qwen2.5-VL}}
& \multirow{2}{*}{3B}
& Original & 0.44 & 0.38 & 0.35 & 0.42 & 0.31 & -- & -- \\
&  & Ours
& 0.51\rlap{\textsuperscript{\cgaphl{+}{0.07}}} & 0.49\rlap{\textsuperscript{\cgaphl{+}{0.11}}} & 0.45\rlap{\textsuperscript{\cgaphl{+}{0.10}}} & 0.49\rlap{\textsuperscript{\cgaphl{+}{0.07}}} & 0.46\rlap{\textsuperscript{\cgaphl{+}{0.15}}} & \textbf{\textcolor{nicergreen}{+0.10}} & \textbf{\textcolor{nicergreen}{+26.3\%}} \\
\cmidrule(lr){2-10}
& \multirow{2}{*}{7B}
& Original & 0.54 & 0.54 & 0.54 & 0.53 & 0.53 & -- & -- \\
&  & Ours
& 0.65\rlap{\textsuperscript{\cgaphl{+}{0.11}}} & 0.67\rlap{\textsuperscript{\cgaphl{+}{0.13}}} & 0.65\rlap{\textsuperscript{\cgaphl{+}{0.11}}} & 0.66\rlap{\textsuperscript{\cgaphl{+}{0.13}}} & 0.64\rlap{\textsuperscript{\cgaphl{+}{0.11}}} & \textbf{\textcolor{nicergreen}{+0.12}} & \textbf{\textcolor{nicergreen}{+22.0\%}} \\
\midrule

\multirow{6}{*}{\textsc{InternVL3}}
& \multirow{2}{*}{2B}
& Original & 0.60 & 0.59 & 0.42 & 0.41 & 0.48 & -- & -- \\
&  & Ours
& 0.62\rlap{\textsuperscript{\cgaphl{+}{0.02}}} & 0.60\rlap{\textsuperscript{\cgaphl{+}{0.01}}} & 0.44\rlap{\textsuperscript{\cgaphl{+}{0.02}}} & 0.42\rlap{\textsuperscript{\cgaphl{+}{0.01}}} & 0.53\rlap{\textsuperscript{\cgaphl{+}{0.05}}} & \textbf{\textcolor{nicergreen}{+0.02}} & \textbf{\textcolor{nicergreen}{+4.4\%}} \\
\cmidrule(lr){2-10}
& \multirow{2}{*}{8B}
& Original & 0.59 & 0.55 & 0.50 & 0.51 & 0.47 & -- & -- \\
&  & Ours
& 0.61\rlap{\textsuperscript{\cgaphl{+}{0.02}}} & 0.61\rlap{\textsuperscript{\cgaphl{+}{0.06}}} & 0.58\rlap{\textsuperscript{\cgaphl{+}{0.08}}} & 0.60\rlap{\textsuperscript{\cgaphl{+}{0.09}}} & 0.55\rlap{\textsuperscript{\cgaphl{+}{0.08}}} & \textbf{\textcolor{nicergreen}{+0.07}} & \textbf{\textcolor{nicergreen}{+12.6\%}} \\
\cmidrule(lr){2-10}
& \multirow{2}{*}{14B}
& Original & 0.61 & 0.51 & 0.42 & 0.40 & 0.40 & -- & -- \\
&  & Ours
& 0.69\rlap{\textsuperscript{\cgaphl{+}{0.08}}} & 0.61\rlap{\textsuperscript{\cgaphl{+}{0.10}}} & 0.62\rlap{\textsuperscript{\cgaphl{+}{0.20}}} & 0.59\rlap{\textsuperscript{\cgaphl{+}{0.19}}} & 0.58\rlap{\textsuperscript{\cgaphl{+}{0.18}}} & \textbf{\textcolor{nicergreen}{+0.15}} & \textbf{\textcolor{nicergreen}{+32.1\%}} \\
\midrule

\multicolumn{10}{@{}c@{}}{\textbf{Close-source VLMs}} \\
\midrule

\textsc{GPT-4o} & -- & API & 0.61 & 0.62 & 0.56 & 0.49 & 0.47 & -- & -- \\
\textsc{Gemini 2.0 Flash} & -- & API & 0.62 & 0.61 & 0.52 & 0.45 & 0.44 & -- & -- \\
\textsc{Claude 3.5 Sonnet} & -- & API & 0.56 & 0.53 & 0.48 & 0.43 & 0.42 & -- & -- \\
\bottomrule
\end{tabularx}
\end{table*}

\begin{table*}[t]
\centering
\caption{\textbf{Performance of vision--language models on regression tasks.} "Ours" refers to applying CADE to the same base model, where per-test gains are highlighted with green.}
\label{tab:regression-results}
\small
\setlength{\tabcolsep}{6pt}
\renewcommand{\arraystretch}{1.15}

\begin{tabularx}{\textwidth}{@{} l c c *{4}{>{\centering\arraybackslash}X} @{}}
\toprule
\textbf{Model} & \textbf{Params} & \textbf{Setting} &
\textbf{MAE}\,$\downarrow$ & \textbf{Rel. Reduction}\,$\downarrow$ & \textbf{Interval Accuracy}\,$\uparrow$ & \textbf{Rel. Gain}\,$\uparrow$ \\
\midrule

\multicolumn{7}{@{}c@{}}{\textbf{Open-source VLMs}} \\
\midrule

\multirow{4}{*}{\textsc{LLaVA-v1.5}}
& \multirow{2}{*}{7B}
& Original & 10.44 & -- & 0.33 & -- \\
&  & Ours & \hspace{1.858em}9.88\cgaphl{-}{0.56} & \textbf{\textcolor{nicergreen}{-5.36\%}} & \hspace{1.858em}0.35\cgaphl{+}{0.02} & \textbf{\textcolor{nicergreen}{+6.06\%}} \\
\cmidrule(lr){2-7}
& \multirow{2}{*}{13B}
& Original & 10.24 & -- & 0.37 & -- \\
&  & Ours & \hspace{1.858em}9.16\cgaphl{-}{1.08} & \textbf{\textcolor{nicergreen}{-10.55\%}} & \hspace{1.858em}0.43\cgaphl{+}{0.06} & \textbf{\textcolor{nicergreen}{+16.22\%}} \\
\midrule

\multirow{4}{*}{\textsc{InstructBLIP}}
& \multirow{2}{*}{7B}
& Original & 22.59 & -- & 0.11 & -- \\
&  & Ours & \hspace{1.858em}11.45\cgaphl{-}{11.14} & \textbf{\textcolor{nicergreen}{-49.31\%}} & \hspace{1.858em}0.20\cgaphl{+}{0.09} & \textbf{\textcolor{nicergreen}{+81.82\%}} \\
\cmidrule(lr){2-7}
& \multirow{2}{*}{13B}
& Original & 10.80 & -- & 0.33 & -- \\
&  & Ours & \hspace{1.858em}9.71\cgaphl{-}{1.09} & \textbf{\textcolor{nicergreen}{10.00\%}} & \hspace{1.858em}0.37\cgaphl{+}{0.04} & \textbf{\textcolor{nicergreen}{+12.12\%}} \\
\midrule

\multirow{2}{*}{\textsc{LLaVA-Onevision}}
& \multirow{2}{*}{7B}
& Original & 10.01 & -- & 0.20 & -- \\
&  & Ours & \hspace{1.858em}8.93\cgaphl{-}{1.08} & \textbf{\textcolor{nicergreen}{-10.79\%}} & \hspace{1.858em}0.25\cgaphl{+}{0.05} & \textbf{\textcolor{nicergreen}{+25.00\%}} \\
\midrule

\multirow{4}{*}{\textsc{Qwen2.5-VL}}
& \multirow{2}{*}{3B}
& Original & 12.09 & -- & 0.27 & -- \\
&  & Ours & \hspace{1.858em}11.93\cgaphl{-}{0.16} & \textbf{\textcolor{nicergreen}{-1.32\%}} & \hspace{1.858em}0.28\cgaphl{+}{0.01} & \textbf{\textcolor{nicergreen}{+3.70\%}} \\
\cmidrule(lr){2-7}
& \multirow{2}{*}{7B}
& Original & 12.48 & -- & 0.22 & -- \\
&  & Ours & \hspace{1.858em}10.45\cgaphl{-}{2.03} & \textbf{\textcolor{nicergreen}{-16.27\%}} & \hspace{1.858em}0.31\cgaphl{+}{0.09} & \textbf{\textcolor{nicergreen}{+40.91\%}} \\
\midrule

\multirow{6}{*}{\textsc{InternVL3}}
& \multirow{2}{*}{2B}
& Original & 10.33 & -- & 0.29 & -- \\
&  & Ours & \hspace{1.858em}9.81\cgaphl{-}{0.52} & \textbf{\textcolor{nicergreen}{-5.03\%}} & \hspace{1.858em}0.37\cgaphl{+}{0.08} & \textbf{\textcolor{nicergreen}{+27.59\%}} \\
\cmidrule(lr){2-7}
& \multirow{2}{*}{8B}
& Original & 19.88 & -- & 0.14 & -- \\
&  & Ours & \hspace{1.858em}14.42\cgaphl{-}{5.46} & \textbf{\textcolor{nicergreen}{-27.46\%}} & \hspace{1.858em}0.20\cgaphl{+}{0.06} & \textbf{\textcolor{nicergreen}{+42.86\%}} \\
\cmidrule(lr){2-7}
& \multirow{2}{*}{14B}
& Original & 24.01 & -- & 0.19 & -- \\
&  & Ours & \hspace{1.858em}11.60\cgaphl{-}{12.41} & \textbf{\textcolor{nicergreen}{-51.69\%}} & \hspace{1.858em}0.25\cgaphl{+}{0.06} & \textbf{\textcolor{nicergreen}{+31.58\%}} \\
\midrule

\multicolumn{7}{@{}c@{}}{\textbf{Close-source VLMs}} \\
\midrule

\textsc{GPT-4o} & -- & API & 11.25 & -- & 0.25 & -- \\
\textsc{Gemini 2.0 Flash} & -- & API & 12.35 & -- & 0.12 & -- \\
\textsc{Claude 3.5 Sonnet} & -- & API & 11.46 & -- & 0.15 & -- \\

\bottomrule
\end{tabularx}
\end{table*}

\vspace{-6pt}
\paragraph{Main benchmark results: MCQ.}
The original MCQ results reveal two clear patterns. First, most open-source VLMs show limited robustness when structured context becomes weaker, with performance generally degrading from stronger-context settings to weaker-context ones. Second, there remains a noticeable gap between weaker open-source models and stronger open- or closed-source systems, suggesting that SDG-oriented multiple-choice reasoning is still challenging even for relatively capable VLMs. Overall, the original results indicate substantial sensitivity to context availability and incomplete generalization under reduced guidance.

\vspace{-4pt}
\paragraph{CADE results: MCQ.}
CADE consistently improves MCQ performance across architectures and parameter scales, and the improvements are especially evident in the harder weak-context tests. This suggests that CADE does not merely raise performance in already favorable settings, but more importantly improves robustness when explicit structured evidence is partially removed or entirely absent. Another notable trend is that the gains are broader for weaker and mid-tier models, while still remaining stable for stronger ones, showing that CADE is effective both as a recovery mechanism for more biased models and as a refinement mechanism for already stronger baselines.

\vspace{-4pt}
\paragraph{Main benchmark results: Regression.}
The original regression results show that continuous-value prediction is also difficult for current VLMs. In general, the models exhibit relatively large estimation errors together with low interval-level accuracy, indicating that they struggle not only with precise value prediction but also with coarse range alignment. Similar to MCQ, the original regression setting reflects clear room for improvement rather than near-saturated performance.

\vspace{-4pt}
\paragraph{CADE results: Regression.}
CADE brings consistent improvements in the regression setting by reducing prediction error and simultaneously improving interval-level correctness. This joint improvement is important because it indicates that CADE is not simply shifting outputs in a way that benefits one metric while hurting another; instead, it improves overall prediction quality. The gains are particularly pronounced for models with weaker original regression behavior, while stronger models still benefit in a stable manner. This mirrors the MCQ findings and supports the view that CADE improves model calibration and robustness across task formats.

\vspace{-4pt}
\paragraph{Overall discussion.}
Taken together, the detailed results show that the original models are consistently vulnerable to SDG-oriented bias and context sensitivity in both discrete and continuous prediction settings. CADE addresses these issues in a broad and stable way: it improves both qualitative and quantitative outcomes without requiring model retraining significantly, and particularly strengthens robustness under weak-context conditions.

\section{Additional Analysis and Discussions}
\label{sec:additional_analysis}

\subsection{Analysis on SDG Bias and CADE's Relation to Contrastive Debiasing}
\label{sec:sdg_vs_general_debiasing}

Debiasing SDG-related predictions is a specific instance of debiasing in general, but it differs in both the source and manifestation of bias. In many general debiasing settings, bias is often associated with socially salient correlations or shortcut features learned from data. In contrast, SDG prediction requires models to infer latent development conditions from noisy, region-dependent multi-modal proxies, such as satellite imagery, street-view imagery, and socio-economic context. When evidence is weak or ambiguous, models therefore tend to rely on generic development priors rather than grounded evidence.

This difference also leads to distinctive bias patterns. As revealed by SDGBiasBench, SDG bias appears as \textbf{pillar-conditioned directional defaults} under Q-only inputs and \textbf{modality imbalance} when structured context dominates visual evidence. Accordingly, debiasing SDG biases requires not only suppressing prior-driven predictions, but also correcting how textual and visual evidence are balanced. CADE is designed for these task-specific challenges: it penalizes unsupported Q-only defaults and adaptively reweights context-dominated logits when image--context disagreement is large. Thus, while CADE follows the general principle of contrastive debiasing, its formulation is tailored to the distinctive structure of SDG monitoring tasks.

\subsection{Analysis on Performance Gains}
We observe cases where the improvement brought by CADE is only incremental. This behavior is expected for contrast-style debiasing methods.
Specifically, for each instance $i$ with finite candidate set $O_i$ (\eg $\{A,B,C\}$), each view $T \in \{T_q, T_{ctx}, T_{img}, T_{full}\}$ produces per-option logits $z_i^{(T)}$, which are converted by softmax into a candidate distribution $p_{i,\cdot}^{(T)}$ over $O_i$. Since contrastive methods operate by comparing two such distributions, their corrective effect depends on the divergence between the contrasted views. When the full-view prediction $p_{i,\cdot}^{(T_{full})}$ is already very close to the text-prior distribution $p_{i,\cdot}^{(T_{ctx})}$, for example when the question prior happens to align with the ground truth, the resulting contrast signal is necessarily weak. In that regime, the correction magnitude becomes small and more susceptible to noise, so the final decoding often changes only slightly and may yield only marginal gains. This interpretation is consistent with the analysis in IBD~\cite{zhu2025ibd}, which similarly notes that small divergence between contrasted predictions leads to limited debiasing effect. More broadly, it also aligns with the same failure mode discussed in earlier contrastive decoding literature, where highly similar contrasted distributions provide only weak or noisy optimization signals~\cite{li2023contrastive,chuang2309decoding}.

\subsection{Analysis on MCQA Selection Bias}
\label{sec:mcqa_selection_bias}

Recent studies show that LVLMs may exhibit MCQA selection bias, where predictions are affected by option positions or option tokens rather than only by question semantics~\citep{janghorbani2023multi, weng2024images, atabuzzaman2025benchmarking}. This issue is particularly relevant to our Q-only analysis, since the model receives no visual or structured evidence. To separate SDG-oriented directional defaults from general MCQA formatting artifacts, we conduct an option-perturbation control experiment on \textsc{InstructBLIP}-7B.

Specifically, we test two types of perturbations: changing the option order and changing the option-token format. After each perturbation, we map the selected answer back to its semantic direction: optimistic, conservative, or pessimistic. As shown in Table~\ref{tab:mcqa_perturbation}, MCQA formatting does affect the exact distribution, especially under option-order changes. However, the effect is limited: across both order and token perturbations, the relative tendency remains consistent as optimistic $>$ pessimistic $>$ conservative. This suggests that the observed Q-only directional default is not solely caused by option-position or option-token artifacts.

Importantly, this control does not change the interpretation of CADE. CADE uses Q-only only as one bias probe, and also relies on CTX+Q and image-present views to penalize context dominance and modality imbalance.

\begin{table*}[t]
\centering
\caption{\textbf{Option-perturbation control experiment on \textsc{InstructBLIP}-7B under the Q-only setting.} Predictions are mapped back to semantic directions after perturbation. Values are percentages.}
\label{tab:mcqa_perturbation}
\small
\begin{tabular}{llccccc}
\toprule
Perturbation type & Perturbed format & Original format & Optimistic & Conservative & Pessimistic \\
\midrule
Original & A/B/C & A/B/C & 56.2 & 20.5 & 23.3 \\
Reverse order & C/B/A & A/B/C & 52.8 & 22.1 & 25.1 \\
Random order & Random permutation & A/B/C & 57.4 & 19.8 & 22.8 \\
\midrule
Lowercase token & a/b/c & A/B/C & 53.9 & 22.1 & 24.0 \\
Numeric token & 1/2/3 & A/B/C & 58.6 & 18.9 & 22.5 \\
\bottomrule
\end{tabular}
\end{table*}

\subsection{Analysis on Efficiency--Effectiveness Tradeoff}
\label{sec:efficiency_tradeoff}
CADE requires four forward passes at inference time, but this cost remains moderate in context. Contrastive debiasing inherently requires multiple predictions, with at least two passes already needed to compare the original and bias-related signals. In CADE, two of the four passes are text-only and are much faster than multi-modal inference. More importantly, this limited overhead yields substantially stronger results than competing methods (as shown in~\Cref{subsec:exp_sota_compare}) while remaining far cheaper and easier to deploy than fine-tuning-based alternatives, which require additional optimization time, GPU resources, and checkpoint management. Overall, CADE offers a favorable efficiency--effectiveness tradeoff in practice.

To make this cost explicit, we report the average per-question runtime on A100 GPUs in~\Cref{tab:runtime_cost} and per-view runtime breakdown analysis in~\Cref{tab:runtime_breakdown}.

\begin{table}[t]
\centering
\caption{\textbf{Inference runtime comparison on A100 GPUs.}}
\label{tab:runtime_cost}
\small
\setlength{\tabcolsep}{4pt}
\begin{tabular}{@{}lcc@{}}
\toprule
Method & \shortstack{GPU-sec\\ / question} & \shortstack{Relative\\ cost} \\
\midrule
Full inference & 0.144 & 1.00$\times$ \\
CADE & 0.324 & 2.25$\times$ \\
\bottomrule
\end{tabular}
\end{table}

\begin{table}[t]
\centering
\caption{\textbf{Approximate per-view runtime breakdown of CADE.}}
\label{tab:runtime_breakdown}
\small
\begin{tabular}{lc}
\toprule
View & Runtime proportion \\
\midrule
Full & 40\% \\
IMG+Q & 40\% \\
CTX+Q & 10\% \\
Q-only & 10\% \\
\midrule
CADE total & 100\% \\
\bottomrule
\end{tabular}
\end{table}

\section{Licenses and Terms of Use}
SDGBiasBench is a derivative benchmark built on the Demographic and Health Surveys (DHS)~\citep{corsi2012demographic} and \emph{DHS-based} subset of \textsc{SustainBench}~\citep{yeh2021sustainbench}, augmented with our human-authored reasoning questions and expert-labeled answers. We release our added benchmark artifacts (questions, answer keys, splits, and evaluation code) under the CC BY-SA 4.0 license, consistent with \textsc{SustainBench}. The underlying modalities follow the original \textsc{SustainBench} data licenses/terms: Landsat, DMSP, NAIP, and VIIRS imagery are public domain; Sentinel-2 imagery is under CC BY-SA 3.0 IGO; Sentinel-1 and MODIS imagery are freely reusable under their respective open access terms; and PlanetScope and Mapillary street-level imagery (when used) are provided under CC BY-SA 4.0 as packaged in \textsc{SustainBench}. We do not redistribute any DHS micro-level survey records; all DHS-derived labels we use are aggregated at the cluster level and comply with the DHS Program Terms of Use.

\section{Intended Use and Access Conditions}
Our benchmark is intended for non-commercial research on multi-modal reasoning and bias diagnosis in SDG-related settings.
We use existing artifacts (\eg pretrained VLMs and publicly available imagery/context sources) in ways consistent with their stated intended use and license/terms, and we do not attempt to recover personal data or identify individuals.
Any derivatives we release are provided for research purposes only and inherit upstream restrictions where applicable; users must comply with original licenses/ToS for any underlying data and models.
We do not endorse deployment of models trained/evaluated on this benchmark for high-stakes decision-making without additional validation and governance.

\section{Information on Use of AI Assistants}
We used AI assistants in a limited capacity to support manuscript preparation, such as refining wording, improving clarity, and checking grammar and formatting. All technical content (\eg ideas, methods, experiments, results, and conclusions) was produced and verified by the authors, who take full responsibility for the final paper.

\end{document}